\documentclass[11pt]{article}

\usepackage[preprint]{acl}

\usepackage{times}
\usepackage{latexsym}

\usepackage[T1]{fontenc}
\usepackage[utf8]{inputenc}

\usepackage{inconsolata}

\usepackage{graphicx}

\usepackage{hyperref}       % hyperlinks
\usepackage{url}            % simple URL typesetting
\usepackage{booktabs}       % professional-quality tables
\usepackage{amsfonts}       % blackboard math symbols
\usepackage{nicefrac}       % compact symbols for 1/2, etc.
\usepackage{microtype}      % microtypography
\usepackage{xcolor}         % colors
\usepackage[textsize=tiny]{todonotes}
\usepackage{color,soul}
\usepackage{subcaption}
\usepackage{multirow}
\usepackage[most]{tcolorbox}
\usepackage{xcolor}        % for green-shaded think blocks (\colorbox)
\usepackage[table]{xcolor} % rowcolor for header strips in figures
\usepackage{enumitem}      % for list spacing
\usepackage{amsmath} 
\usepackage{tabularx}
\usepackage{adjustbox,array}
\usepackage{cleveref}
\usepackage{wrapfig}
\usepackage{xspace}
\usepackage{placeins}   % \FloatBarrier, used to keep appendix tables with their subsection

\setlist[itemize]{leftmargin=1.2em,topsep=3pt,itemsep=3pt,parsep=0pt}
\colorlet{soulblue}{blue!20}

\newcommand{\ours}{\textsc{TimeThink}\xspace}
\newcommand{\oursSFT}{\textsc{TimeThink}~(SFT)\xspace}
\newcommand{\oursRL}{\textsc{TimeThink}~(RL)\xspace}
\definecolor{gain}{HTML}{1A7F37}
\definecolor{loss}{HTML}{C24D4D}

\newcommand{\gain}[1]{\textcolor{gain}{\textbf{#1}}}
\newcommand{\loss}[1]{\textcolor{loss}{\textbf{#1}}}

\newcommand{\metric}{metricX}
\newcommand{\metricA}{metricA}
\newcommand{\metricB}{metricB}
\newcommand{\metricAnchor}{metricAnchor}

\usepackage{pgfplots}
\pgfplotsset{compat=1.18}
\usetikzlibrary{patterns,decorations.pathreplacing,calc}

\title{\ours: Eliciting Compositional Reasoning in Timeseries Large Language Models}

\author{
  \textbf{Sudarshan Regmi}, \textbf{Arvind Pillai}, \textbf{Yu Yvonne Wu}, \textbf{Yuliang Chen},
\\
  \textbf{Bibek Panthi}, \textbf{Tess Z. Griffin}, \textbf{Michael V. Heinz},
\\
  \textbf{Lisa Marsch}, \textbf{Nicholas C. Jacobson}, \textbf{Andrew Campbell}
\\
\\
  Dartmouth College
\\
  \small{
    \textbf{Correspondence:} \href{mailto:sudarshan.regmi.gr@dartmouth.edu}{sudarshan.regmi.gr@dartmouth.edu}
  }
}

\begin{document}
\maketitle
\begin{abstract}
Timeseries multimodal large language models (TS-MLLMs) have recently begun leveraging the reasoning capabilities of large language models (LLMs) for question-answering tasks. However, these models often fail to capture dynamic temporal patterns, providing only implicit reasoning that lacks the underlying explanations critical for high-stakes applications like healthcare. While reinforcement learning (RL)-based timeseries language models aim to address this, they often fall short because they are trained on narrow, in-distribution data and struggle with out-of-distribution compositional questions. To address these challenges, we present \textbf{\ours{}}, a synthetic framework for eliciting compositional timeseries reasoning. Core timeseries primitives (e.g., trend, seasonality) are domain-independent and can be deterministically generated. Guided by this premise, \ours{} first designs a synthetic data generator that produces \textit{atomic} and \textit{composite} question-answer pairs, providing objective ground truth with reasoning traces. Building on this framework, \ours{} employs a reinforcement learning with verifiable rewards (RLVR) training strategy that encourages explicit reasoning. Unlike template-reliant methods, this approach enables the model to learn the underlying logic of composition rather than simply imitating traces. Extensive experiments show that \ours{}, trained only on synthetic data, significantly outperforms strong baselines on both synthetic and real-world benchmarks.\footnote{Code: \url{https://github.com/sudarshanregmi/timethink}}
\end{abstract}
\section{Introduction}
Timeseries data are ubiquitous in real-world systems, capturing temporal dynamics across healthcare, industrial operations, and finance. Recent timeseries multimodal large language models (TS-MLLMs) adapt large language models (LLMs) to numerical sequences and have shown promising results on detection and question-answering tasks~\citep{xie2024chatts, zhang2025sensorlm}. Existing TS-MLLMs differ mainly in how the numerical signal is encoded for the language model. Some convert the series to raw text or visual plots so that an LLM or vision-language model (VLM) can interpret it directly~\citep{yoon2024my, kim2024health}. Others embed the series natively into the language space, training the projection on paired timeseries and language data~\citep{xie2024chatts, xu2025lens, zhang2025sensorlm}. Across these designs, however, the model is supervised only on the final answer. Temporal patterns such as trends, seasonality and anomalies are absorbed implicitly, and the reasoning behind a prediction is left invisible~\citep{li2026hearts}. This matters because the action taken downstream of a timeseries model usually depends on which pattern triggered the prediction, not just on the prediction itself. For example, when an equipment failure occurs in rotating machinery, a model may correctly raise an alarm from vibration, current, and temperature streams. But it does not reveal whether the alarm stems from sustained motor load, rising bearing temperature, or a transient vibration spike. This distinction is critical for routing maintenance to the right component. The same holds in healthcare and other high-stakes settings, where actions require justified diagnoses.

These settings demand not only correct answers but the reasoning behind them. To this end, a recent line of work uses reinforcement learning with verifiable rewards to elicit reasoning traces~\citep{Heetal2026, guan2026timeomni, streasoner}. Models such as TimeOmni-1~\citep{guan2026timeomni} and STReasoner~\citep{streasoner} are optimized against outcome-based rewards: typically exact match on multiple-choice answers or error on a forecast. In practice, however, the bulk of capability in these works is installed by supervised fine-tuning (SFT) on reasoning templates, with reinforcement learning (RL) serving as a refinement pass that sharpens format compliance and offers modest accuracy gains. Together, this recipe makes the reasoning legible at inference time and improves accuracy on the targeted tasks.

Two limitations follow from this design. First, because the reasoning templates are crafted or distilled per task, the resulting models are tied to the narrow band of temporal patterns covered by their training distribution (a single forecasting target, a fixed sensor domain, or a closed set of question types) and might not transfer to timeseries whose properties were unseen at training time. Second, real-world questions rarely turn on a single temporal phenomenon. They typically require composing several at once: locating an anomaly within a seasonal trend, or correlating two variables under a non-stationary regime. When a question spans multiple patterns composed in novel ways, many TS-MLLMs lose the thread and fall back to the heuristics induced by their reasoning templates.

To address these gaps, we propose \ours{}, a synthetic framework for eliciting compositional timeseries reasoning. The framework begins with a synthetic data generator that combines domain-invariant \emph{atomic} primitives with a separate set of \emph{sub-skills} to form \emph{composite} QAs. For a small subset of examples, we use an LLM to diversify the wording and complexity of the questions and answers while preserving the programmatically generated reasoning structure and verified ground truth. The resulting mixture of atomic and composite QAs is then used in two sequential training stages, SFT followed by reinforcement learning with verifiable rewards (RLVR) to enable explicit reasoning. RLVR is driven by question-aware rewards that decompose by answer type (binary, categorical, numerical, and set overlap), thereby providing denser learning signals. We evaluate \ours{} on both synthetic and real-world question-answering benchmarks, which highlight the efficacy of a synthetic-only regime. Our contributions:

\begin{itemize}
    \item \textbf{\ours-framework:} We develop a programmatic generator for domain-independent timeseries primitives that produces deterministic question-aware ground truth, making RLVR tractable at scale. We create evaluation sets containing varying compositions of timeseries primitives. We plan to release all code, data, and models to encourage open-source research.
    \item \textbf{Timeseries compositional reasoning:} We study compositional reasoning over timeseries, where answering a query requires composing learned \emph{atomic primitives}. To the best of our knowledge, we present the first work to train such reasoning explicitly with synthetic verifiable rewards in TS-MLLMs.    
    \item \textbf{Extensive analysis:} We evaluate on diverse synthetic and real-world datasets across six tasks. Results on compositional questions demonstrate that \ours{} outperforms existing timeseries models, and further analysis shows \oursRL{} generalizes to extended temporal horizons.
\end{itemize}
\section{Related Work}

\paragraph{TS-MLLMs without explicit reasoning.}
Existing TS-MLLMs differ primarily in how the numerical signal is presented to the language model. Early methods convert timeseries into strings for direct LLM processing~\citep{gruver2023large, xue2023promptcast, kim2024health}, which suffer from token exhaustion and degraded fidelity over long sequences~\citep{pillai2025time2lang, spathis2024first}. To improve efficiency, alternative strategies use visual prompting via vision-language models~\citep{yoon2024my}, but these often pick up on surface visual patterns rather than the underlying temporal dynamics~\citep{mcq2}. A more recent line treats timeseries as a native modality, projecting numerical patches directly into the LLM embedding space and aligning the two via paired data~\citep{jin2023time, xie2024chatts, xu2025lens, zhang2025sensorlm}. In all three cases, models are often supervised only on final answers. Temporal patterns are absorbed implicitly, so the model cannot expose or audit the reasoning behind a prediction. Recent benchmarks show that even frontier LLMs struggle with synthetic timeseries questions beyond basic pattern recognition~\citep{cai2024timeseriesexam}, with performance deteriorating further on multi-step temporal reasoning tasks~\citep{li2026hearts}.

\paragraph{Explicit reasoning over timeseries and its limits.} A recent work~\citep{potosnak2024} shows deep forecasters can compose temporal structure under distribution shift, but that composition is implicit and never trained for. More recently, researchers have started to explore explicit reasoning over timeseries, following the broader RL-for-reasoning paradigm~\citep{guo2025deepseek, shao2024deepseekmath}. One line supervises on annotated chain-of-thought traces along with the final answer (SFT-style), as in SenTSR-Bench~\citep{Heetal2026} and LLaTiSA~\citep{ding2026llatisa}. Another uses reinforcement learning with task-specific rewards to elicit thinking traces during training, including STReasoner~\citep{streasoner} and TimeOmni-1~\citep{guan2026timeomni}. A parallel line applies RL to temporal reasoning over events and dates~\citep{liu2025timer1}. Two limitations curb both lines. First, they are typically trained on relatively narrow task distributions, which can limit generalization to out-of-distribution timeseries. Second, they can struggle with questions that require composing multiple temporal patterns. What remains missing is explicit reasoning over core temporal primitives, such as trend, correlation, and seasonality with verifiable rewards. Our work targets this gap.
\section{\ours}

\begin{figure*}
    \centering
    \includegraphics[width=0.925\linewidth]{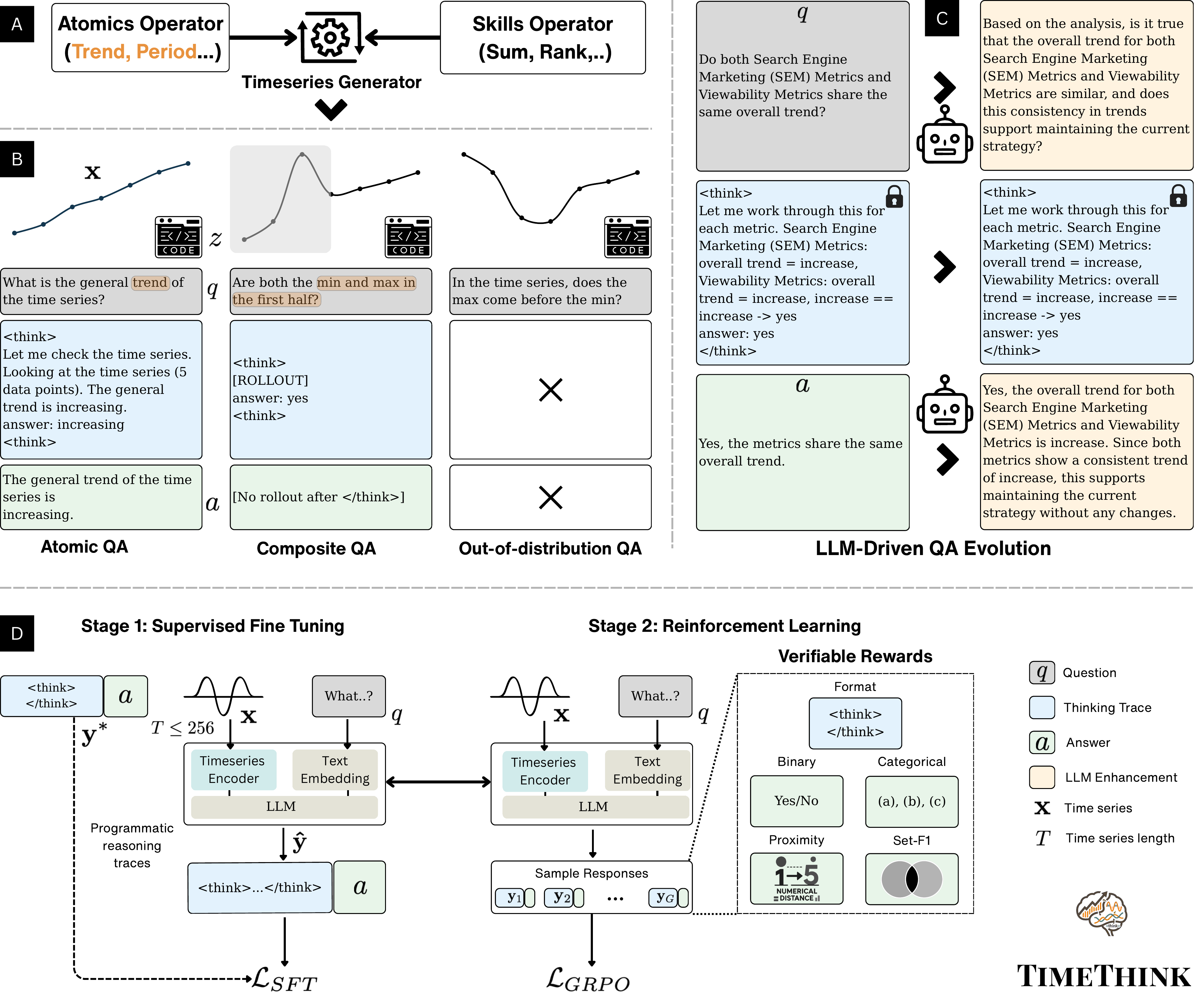}
    \caption{\textbf{Overview of \ours}.
    \textbf{(A, B) Synthetic Data Generation:} a programmatic generator builds Atomic QA targeting single primitives, Composite QA chaining multiple operators, and \textit{out-of-distribution (OOD) QA} over unseen task axes. \textbf{(C) LLM-Driven QA Evolution:} templated QAs are evolved into more complex, diverse variants, with the reasoning trace (\texttt{<think>} block) preserved so that responses stay grounded in timeseries properties. \textbf{(D) Training Stage Optimization:} \textit{Stage 1 (SFT)} fine-tunes on programmatic reasoning traces at an anchor length of $T\le256$; \textit{Stage 2 (RLVR)} applies GRPO to incentivize autonomous composition and extend the horizon to $T\le768$, using a verifiable reward ($r_{\mathrm{fmt}} + r_{\mathrm{ans}}$) over four answer categories (binary, proximity, categorical, Set-$F_1$) besides the format reward.}
    \label{fig:overview}
\end{figure*}

\paragraph{Preliminaries.} We consider the task of question answering over multivariate timeseries. Let $\mathbf{x} = (x_1,\dots,x_T) \in \mathbb{R}^{T \times d}$ denote a timeseries of length $T$, where each observation $x_t \in \mathbb{R}^d$ corresponds to a univariate ($d=1$) or multivariate ($d>1$) sequence. Given a question $q \in \mathcal{Q}$ about $\mathbf{x}$, the model generates an autoregressive response $\mathbf{y} = (y_1,\dots,y_{|\mathbf{y}|})$ comprising a reasoning trace and a final answer. To facilitate the independent verification of response structure and grounding, we employ a structured output format: \texttt{<think>}$\cdots$\texttt{answer: \{value\}}\texttt{</think>}\allowbreak$\langle$\texttt{post-think response}$\rangle$, enabling verification of response format and answer correctness. 
The conditional distribution $\pi_\theta(\mathbf{y}\mid \mathbf{x}, q)$ over output tokens is defined as:

\begin{equation}\pi_\theta(\mathbf{y}\mid \mathbf{x}, q) =\prod_{n=1}^{|\mathbf{y}|} \pi_\theta(y_n \mid \mathbf{x}, q, y_{<n}),
\end{equation}

where $\theta$ represents the model parameters. It underlies both supervised fine-tuning and reinforcement learning with verifiable rewards.

\noindent\textbf{Overview.} \ours{} introduces a synthetic framework for eliciting compositional timeseries reasoning by incorporating (a) synthetic data generation and (b) training with SFT followed by RLVR. Data generation synthesizes and evolves atomic and composite QAs using an LLM, while training is designed to incentivize the model to compose learned atomic operations. We describe \ours{}'s components below.

\subsection{Synthetic Data Generation} \label{sec:synthetic_data_generation}
Timeseries reasoning has a natural compositional structure. Questions may depend not only on a single temporal pattern but on mixtures of trends, seasonality, and others. Answering them requires identifying primitive temporal facts and composing them through higher-level reasoning operations. To mimic this structure, we design a synthetic QA generator. Unlike open-ended language or visual reasoning, these primitives can be procedurally generated with deterministic intermediate states and exact ground truth. Specifically, to simulate real-world temporal patterns, we design comprehensive temporal QA generators producing \emph{atomic} and \emph{composite} questions. Each composite question in our corpus is built from two kinds of operators. An \emph{atom operator} reads the timeseries directly and returns one fact, for example a maximum value or the position of a peak. The atom operator set $\mathcal{F}_{\mathrm{atom}}$ aggregates over nine common temporal operations: \textsc{Mean}, \textsc{Std}, \textsc{Percentile}, \textsc{ExtVal}, \textsc{ExtPos}, \textsc{EventEnum}, \textsc{SegEnum}, \textsc{Period} and \textsc{TrendClassify}.
A \emph{sub-skill operator} takes one or more facts (from atoms, or from sub-skills) and combines them into a final answer. The \emph{sub-skill set} $\mathcal{F}_{\mathrm{op}}$ applies diverse reasoning operations to atomics: \textsc{Count}, \textsc{Sum}, \textsc{Filter}, \textsc{Locate}, \textsc{Compare}, \textsc{Threshold}, \textsc{Argmax}, \textsc{Rank} and \textsc{Arith}.
In short, atoms tell the model \emph{what to look for} and sub-skills \emph{how to combine what it found}.

\noindent\textbf{Atomic QA.} An atomic QA instance is a single atom application: $a = f(\mathbf{x})$ with $f \in \mathcal{F}_{\mathrm{atom}}$. We refer to the output $f(\mathbf{x})$ of an atom operator as an \emph{atomic fact}. 
For example, $a = \textsc{ExtPos}(\mathbf{x})$ is answer to \emph{``At which timestep does $\mathbf{x}$ reach its maximum?''}. Atomic QAs teach the primitives thereby aligning the timeseries encoder with LLM backbone that every downstream composition will invoke.

\noindent\textbf{Composite QA.} Real-world questions about timeseries are rarely about one pattern in isolation. A practitioner usually asks about a mix of patterns at once, for instance comparing two segments, locating an extremum within a trend, or checking whether two signals agree on their overall shape. Composite QAs mimic this structure: each question is built by reading several atomic facts from the timeseries and then combining them with one or more sub-skills. Concretely, a composite QA uses $m \geq 1$ atom operators $f_i \in \mathcal{F}_{\mathrm{atom}}$, each applied to $\mathbf{x}$ or to a sub-series of $\mathbf{x}$, to extract facts, and then $K \geq 1$ sub-skill operators $g_j \in \mathcal{F}_{\mathrm{op}}$ to combine those facts into the final answer $a$:
\[
\underbrace{f_1(\mathbf{x}),\ \ldots,\ f_m(\mathbf{x})}_{m\ \text{atomic facts}}
\;\xrightarrow{\;g_1,\, g_2,\, \ldots,\, g_K\;}\;
a.
\]
For example, the question \emph{``Is the first half of $\mathbf{x}$ on average higher than the second half?''} reflects a common real-world question about a timeseries. We build it by reading two atomic facts, the means of the two halves, and then combining them with a comparison: $\textsc{Mean}(I_1),\ \textsc{Mean}(I_2)\xrightarrow{\textsc{Compare}} a$, where $I_1 = (x_1,\dots,x_{\lfloor T/2 \rfloor})$ and $I_2 = (x_{\lfloor T/2 \rfloor + 1},\dots,x_T)$ denote the first and second halves of $\mathbf{x}$. Every composite question thus has a known program $a = (g_K \circ \cdots \circ g_1)(f_1(\mathbf{x}),\ldots,f_m(\mathbf{x}))$. By \emph{\textbf{compositional reasoning}}, we mean answering questions that require combining multiple atomic facts. After building the QA structures above, we partition composite QA families into three splits, $\mathcal{Q}_{\mathrm{D}}$, $\mathcal{Q}_{\mathrm{R}}$, and $\mathcal{Q}_{\mathrm{ood}}$, according to the supervision each family receives. Our training pipeline has two stages (detailed in \Cref{sec:method_training}): SFT on worked decompositions, followed by RLVR. The three splits differ in how much of this pipeline each family participates in:

\noindent\textbf{(a) Comp-D (CoT-demonstrated)}: a chain-of-thought (CoT) decomposition is shown during SFT, and the same composite additionally appears in the RL pool. \textbf{(b) Comp-R (reward-only)}: the composite appears only in the RL pool, so the policy is expected to solve it from the verifiable outcome reward alone. \textbf{(c) OOD (held-out)}: the composite is held out from both training stages and evaluated only at test time, which measures whether compositional reasoning learned in (a) and (b) transfers to compositions the policy never saw.

A composite QA answer $a$ can admit multiple equivalent decompositions, so SFT on a single CoT decomposition risks overfitting to a stylistic template, while omitting atomic QAs during RL risks primitive-skill regression and rote memorization of composites. We therefore mix atomic and composite QAs in both the SFT and RL phases. The complete enumeration of every QA family is provided in \Cref{app:atomic}--\labelcref{app:ood}.

\subsubsection{LLM-driven QA Evolution}
\label{sec:tsevol}
The generator produces questions from a fixed set of templates. Training only on these risks memorization of form rather than learning timeseries structure. To prevent this, we apply an LLM-driven \emph{question evolution} step to a small portion of training data to diversify question and answer framing (Figure \ref{fig:overview}C). Following \citet{xu2024wizardlm}, we evolve our atomic and composite templated QAs (containing explicit reasoning breakdown) into LLM-enhanced QAs. Specifically, we condition the LLM on the original question, reasoning block, and answer, and prompt it to generate a more complex question-answer pair while preserving the same reasoning structure. The evolution strategy is selected automatically based on the seed example \citep{xu2024wizardlm}. Unlike prior approaches such as ChatTS~\citep{xie2024chatts}, our framework preserves explicit reasoning grounding between the question and the final answer. This keeps the model grounded in fundamental timeseries properties rather than shortcuts, and robust to elaborate rephrasings. We additionally use question paraphrasing as a lightweight augmentation, retaining the same reasoning block and final answer.

\subsection{Training} \label{sec:method_training}
Following ChatTS~\citep{xie2024chatts}, we adopt timeseries encoder-fused LLM and value-preserved timeseries normalization to train the model. The data generator in \Cref{sec:synthetic_data_generation} gives us a test bed for compositional timeseries reasoning: every QA in this test bed has a known reasoning chain (the atoms and sub-skills that lead to the answer) and a verifiable final answer, and the Comp-D/Comp-R/OOD splits separate compositions that are demonstrated, rewarded-only, and held out respectively. As discussed, many existing TS-MLLMs are trained primarily with final-answer supervision, providing no signal for producing a reasoning chain. Therefore, to make the reasoning explicit, we design a two-stage training framework that elicits the reasoning chain.

\noindent\textbf{Supervised Fine-Tuning.} The first stage teaches the model what a valid reasoning chain looks like. We fine-tune on (timeseries, question, target) triples where the target is a worked chain followed by the final answer, so the model learns to emit both, in order, when prompted with a question. In the supervised stage, we train on synthetic demonstrations
$\mathcal{D}_{\mathrm{SFT}}=\{(\mathbf{x}^{(i)}, q^{(i)}, \mathbf{y}^{*(i)}) \}_{i=1}^{N_{\mathrm{SFT}}}$
where each reference target $\mathbf{y}^{*(i)}$ is a worked reasoning trace followed by the final answer. The SFT objective
maximizes the log-likelihood of the reference output:
\begin{equation}
\mathcal{L}_{\mathrm{SFT}}(\theta)
= - \mathbb{E}_{\mathcal{D}_{\mathrm{SFT}}}
\Big[
\sum_{n=1}^{|\mathbf{y}^*|}
\log \pi_\theta(y_n^* \mid \mathbf{x}, q, y_{<n}^*)
\Big].
\label{eq:sft}
\end{equation}

\noindent\textbf{Reinforcement Learning with Verifiable Rewards.} 
The second stage refines the model's reasoning under a verifiable reward. The model generates its own chain on each example and is scored only on whether its final answer is correct and whether the response follows the required format. Because the reward checks only the answer, the model is free to discover chains that go beyond the ones it imitated during SFT. Given a timeseries and a question $(\mathbf{x}, q)$, we sample a group of $G$ responses $\{\mathbf{y}_g\}_{g=1}^{G} \sim \pi_{\theta_{\mathrm{old}}}(\cdot \mid \mathbf{x}, q)$. Each response is evaluated using a verifiable reward function $r(\mathbf{x}, q, \mathbf{y})$, which decomposes into two additive components: $r(\mathbf{x}, q, \mathbf{y}) = r_{\mathrm{fmt}}(\mathbf{y}) + r_{\mathrm{ans}}(\mathbf{x}, q, \mathbf{y})$, with $r_{\mathrm{fmt}}, r_{\mathrm{ans}} \in [0, 1]$ and hence $r   \in [0, 2]$. The first component $r_{\mathrm{fmt}}$ checks structural
compliance for reliable extraction of predicted answer such that the response must contain exactly one
\texttt{<think>...</think>} block whose final line follows the template
\texttt{answer: \{value\}}. The accuracy component, $r_{\mathrm{ans}}$, performs question-aware verification on the extracted value across four distinct categories: (1) \textbf{Binary}: exact match is required for boolean questions; (2) \textbf{Categorical}: exact match is required over a finite semantic label space; (3) \textbf{Proximity}: for numerical queries, a graded reward signal is assigned based on the distance between the predicted value and the ground truth; and (4) \textbf{Set-$F_1$}: partial credit is awarded for enumeration or grouping tasks by computing the $F_1$ overlap between the predicted and gold sets. This reward design eliminates the need for manually annotated reasoning traces while still providing a rich and informative training signal. Ultimately, we optimize the Group Relative Policy Optimization (GRPO) objective~\citep{shao2024deepseekmath}:

\begin{equation}
  \resizebox{0.97\columnwidth}{!}{$\displaystyle
  \max_{\theta}\,
  \mathbb{E}_{(\mathbf{x},q),\,\mathbf{y}_g \sim \pi_{\theta_{\mathrm{old}}}}
  \big[
  \mathcal{L}_{\mathrm{GRPO}}(\theta,\{r(\mathbf{x},q,\mathbf{y}_g)\}_{g=1}^{G})
  \big]
  $},
  \label{eq:grpo}
\end{equation}
which normalizes rewards within each group to obtain relative advantages and updates the policy with a PPO-style clipped surrogate plus a KL penalty against a frozen reference policy $\pi_{\mathrm{ref}}$~\citep{guo2025deepseek}. Here $\pi_{\theta_{\mathrm{old}}}$ denotes the rollout policy held fixed during each PPO update, while $\pi_{\mathrm{ref}}$ is the SFT-stage model frozen throughout RL. We use the implementation provided by the verl library~\citep{sheng2025hybridflow}.
\section{Experiments}
\label{sec:experiments}

\noindent\textbf{Task.} Our evaluation covers six task families: multiple-choice, categorical, numerical, clustering, ranking/enumeration/ordering and description questions.

\noindent\textbf{Datasets.}
We evaluate on the following datasets: (a) TSEvol~\citep{xie2024chatts}: It includes inductive (LLM-as-judge score), causal (accuracy) and deductive (accuracy) questions collected from real-world domains. (b) MCQ2~\citep{mcq2}: It includes counterfactual MCQ questions relating to pair of timeseries. (c) SenTSR-Bench~\citep{Heetal2026}: It includes MCQ questions involving anomaly characterization (\textsc{What Happened}), cause diagnosis (\textsc{How Happened}) and action recommendation (\textsc{Suggested fix}) questions. (d) \ours~bench: See \Cref{app:atomic}--\labelcref{app:ood}.

\noindent\textbf{Metrics.} (a) Accuracy: match to ground truth; (b) Proximity accuracy: computed for numerical tasks as $\max\!\bigl(0,\,1-|\hat e-e|/s\bigr)$, where $\hat e$ is the prediction, $e$ is the ground truth, and the scale $s$ is the sequence length for positional tasks and $\max(|e|,1)$ otherwise; (c) F1: computed for grouping tasks over groups extracted by the LLM judge; (d) Judge-score: computed by the LLM judge (Qwen2.5-72B-Instruct-GPTQ-Int4) depending on the task (ranking, enumeration, ordering, or description).

\noindent\textbf{Baselines.} For general-purpose LLMs, we use Qwen2.5-Instruct-7B \citep{qwen25} and Mistral-Instruct-7B-v0.3~\citep{jiang2023mistral7b}. For text-based timeseries and temporal reasoning models, we include Time-R1-3B \citep{liu2025timer1}, Time-MQA-7B \citep{kong2025time}, and TimeOmni-1 \citep{guan2026timeomni}. For timeseries encoder-based approaches, we compare against ChatTS \citep{xie2024chatts}.

\noindent\textbf{Training details.} We train three variants of Qwen3-8B~\citep{yang2025qwen3technicalreport} augmented with a timeseries encoder~\citep{xie2024chatts} on a single node with 8$\times$NVIDIA H200 GPUs using the verl~\citep{sheng2025hybridflow} framework. The first variant (equivalent to ChatTS~\citep{xie2024chatts}) is trained on answer-only targets over our task taxonomy. The second variant, \oursSFT, adds chain-of-thought supervision. The third variant, \oursRL, initializes from the SFT checkpoint and applies RLVR with GRPO~\citep{shao2024deepseekmath} optimization. We use AdamW with weight decay $0.01$ for all three post-training stages. Both ChatTS and \oursSFT use a learning rate $1\mathrm{e}{-5}$ and effective batch size $192$. GRPO optimization uses learning rate $1\mathrm{e}{-6}$ and effective batch size $256$ with 16 rollouts. The training corpus consists entirely of synthetic data (90K SFT samples and 110K RL samples), with no real-world timeseries data included. There are 7.5k LLM-evolved QAs in the SFT pool, whereas there are 7.5k question-rephrased samples in the RL pool.

\section{Results}
In all tables, \textbf{bold} marks the best score and \underline{underline} the second-best. Cross-architecture comparisons are restricted to sequence length below $256$ due to the cost of text-based models; among TS-native models we use lengths up to $1536$.

\begin{table*}[t]
\centering
\caption{\textbf{Performance on synthetic compositional timeseries QA.} \oursRL{} outperforms all baselines.}
\label{tab:main-results-breakdown}
\begin{adjustbox}{max width=0.85\textwidth,center}
\begin{tabular}{l ccccc | ccc}
\toprule
\multirow{2}{*}{\textbf{Model}} & \multicolumn{5}{c|}{\textbf{Comp-D} \scriptsize(\#3{,}261)} & \multicolumn{3}{c}{\textbf{Comp-R} \scriptsize(\#1{,}948)} \\
\cmidrule(lr){2-6} \cmidrule(lr){7-9}
& Categorical & Numerical & Clustering & Ord./Enum. & {W. Avg} & Categorical & Numerical & {W. Avg} \\
\midrule
Qwen2.5-Instruct-7B     & 0.678 & 0.474 & 0.263 & 0.484 & 0.560             & \underline{0.656} & 0.472             & \underline{0.593} \\
Mistral-7B-Instruct-v0.3 & 0.624 & 0.406 & 0.279 & 0.432 & 0.507             & \underline{0.656} & 0.422             & 0.576             \\
Time-MQA                & 0.538 & 0.364 & 0.189 & 0.374 & 0.438             & 0.550             & 0.371             & 0.488             \\
Time-R1                 & 0.605 & 0.436 & 0.233 & 0.487 & 0.511             & 0.597             & 0.409             & 0.532             \\
TimeOmni-1              & 0.574 & 0.517 & 0.247 & 0.403 & 0.512             & 0.552             & \underline{0.547} & 0.550             \\
\midrule
ChatTS                  & 0.550 & \underline{0.650} & 0.407 & 0.547 & 0.573 & 0.525 & 0.496 & 0.515 \\
\oursSFT                & \underline{0.722} & 0.641 & \underline{0.707} & \underline{0.728} & \underline{0.696} & 0.562 & 0.474 & 0.532 \\
\textbf{\oursRL}        & \textbf{0.771}    & \textbf{0.668}    & \textbf{0.712}    & \textbf{0.750}    & \textbf{0.731}    & \textbf{0.722}    & \textbf{0.754}    &
\textbf{0.733} \\
\bottomrule
\end{tabular}
\end{adjustbox}
\end{table*}

\noindent\textbf{Comp-D vs.\ Comp-R:} The results in \Cref{tab:main-results-breakdown}, evaluated on $\sim$5k test samples, mirror the asymmetry between the two splits. On Comp-D, \oursSFT{} already reaches 0.696, since the demonstrated decompositions cover most of the reasoning the model must produce, and RL adds a further $+3.5$~pp ($0.696 \to 0.731$). On Comp-R, \oursSFT{} plateaus at 0.532, consistent with the difficulty of transferring SFT demonstrations to composite families absent from the SFT pool. RL lifts this to 0.733, a \textbf{$+20.1$}~pp gain. This gain shows that reward-only training substantially improves performance on composite families that lack SFT reasoning demonstrations. To separate the effect of RL from additional exposure to Comp-R examples, we compare RL against rejection fine-tuning from the same checkpoint using the same 10K prompts (\Cref{tab:additional_sft_rl}). RL improves Comp-R from 0.528 to 0.597 compared with 0.567 for additional SFT. These controlled results suggest that the gains are not explained solely by additional training exposure.

\begin{figure}[t]
\centering
\includegraphics[width=\linewidth]{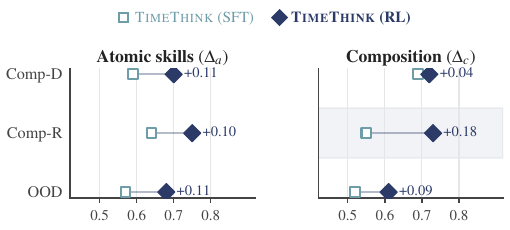}
\caption{RL gains come from both the atomic skills ($\Delta_a$) and the sub-skill that composes them ($\Delta_c$).}
\label{fig:rl_gain}
\end{figure}

\begin{figure}[b]
\centering
\includegraphics[width=\linewidth]{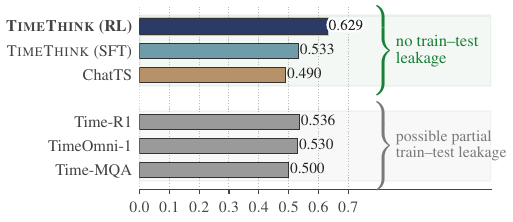}
\nolinenumbers
\caption{Overall OOD performance across tasks. \oursRL{} leads all native-TS baselines.}
\label{fig:ood_performance}
\end{figure}

\noindent\textbf{OOD composition:} A strictly fair OOD comparison applies only to timeseries-native models, whose encoders are trained on a fixed distribution; text-based baselines may have seen similar patterns during pretraining, so their OOD numbers are not directly comparable. \Cref{fig:ood_performance} reports overall OOD compositional performance (mean over numerical, categorical and enumeration) on $\sim$8k test samples. First, \oursRL improves over its SFT counterpart by nearly 10 pp on the OOD split (0.533\,$\to$\,0.629). Second, the \oursSFT model yields only marginal improvements over the ChatTS baseline, suggesting that imitation alone does not transfer under distribution shift. Third, text-based baselines at best match \oursSFT despite Time-MQA and TimeOmni-1 being trained on timeseries data, and Time-R1 on temporal-reasoning data.

\noindent\textbf{Where does compositional ability come from?} A composite question requires both the constituent atomic skill(s) and the \emph{sub-skill(s)} that combine them, so a gain on the overall composition could reflect either. For every
composition, we score its constituent atoms on their dedicated test set and compare the RL gain on atoms ($\Delta_a$) against the gain on the composition ($\Delta_c$) (unweighted means). Three patterns drive the reading of \Cref{fig:rl_gain} and per-composition details are in \Cref{app:comp_breakdown}. RL improves the atomic substrate by the same amount regardless of which compositions sit on top of it, so the variance in $\Delta_c$ is not driven by variance in $\Delta_a$. Atomic propagation alone would predict $\Delta_c \le \Delta_a$ in every row; the Comp-R row falsifies this ($\Delta_c=+0.18 > \Delta_a=+0.10$). With atomic gain held essentially constant, the $+0.18$ on Comp-R against $+0.04$ on Comp-D reflects RL acquiring the sub-skill that combines atoms where SFT did not demonstrate it. The $+0.09$ on OOD additionally indicates that this learned sub-skill transfers to held-out composite questions.

\begin{table}[t]
\centering
\caption{\textbf{Performance on real-world TSEvol datasets.} Trained only on synthetic QA, \oursRL{} achieves the best weighted average.}
\label{tab:real-world-reasoning}
\begin{adjustbox}{max width=\columnwidth,center}
\begin{tabular}{@{}l cccc@{}}
\toprule
Model & Deductive & Causal & Inductive & W. Avg \\
\midrule
Qwen2.5-7B       & \textbf{0.628}    & 0.533             & 0.420             & 0.537 \\
Mistral-7B       & 0.442             & \underline{0.685} & \underline{0.798} & 0.642 \\
Time-MQA         & 0.581             & 0.511             & 0.385             & 0.506 \\
Time-R1          & \underline{0.605} & 0.391             & 0.552             & 0.476 \\
TimeOmni-1       & \textbf{0.628}    & 0.620             & \underline{0.798} & \underline{0.654} \\
\midrule
ChatTS           & 0.442             & 0.630             & 0.747             & 0.602 \\
\oursSFT         & \underline{0.605} & 0.620             & 0.712             & 0.632 \\
\textbf{\oursRL} & \underline{0.605} & \textbf{0.717}    & \textbf{0.835}    & \textbf{0.709} \\
\bottomrule
\end{tabular}
\end{adjustbox}
\end{table}

\noindent\textbf{Real-world evaluation.} On the real-world TSEvol~\citep{xie2024chatts} datasets (\Cref{tab:real-world-reasoning}), \oursRL{} achieves the best results on causal and inductive reasoning as well as on average, improving over \oursSFT{} from 0.632 to 0.709, while text-based baselines such as Time-MQA and Time-R1 underperform general-purpose text baselines. RLVR on basic timeseries skills hence transfers to real-world reasoning tasks.

\section{Ablations and Case Studies}

\begin{figure}[t]
\centering
\includegraphics[width=\linewidth]{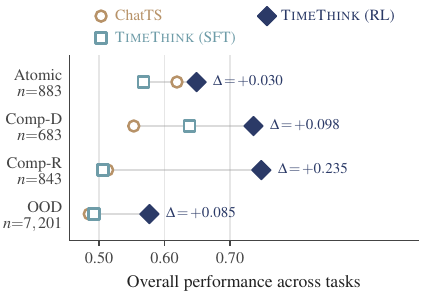}
\caption{Overall performance on unseen-length timeseries ($[769,1536]$ steps). $\Delta$ is \oursRL's gain over the best baseline.}
\label{fig:length-ood-by-category}
\end{figure}

\begin{figure}[b]
\centering
\includegraphics[width=\linewidth]{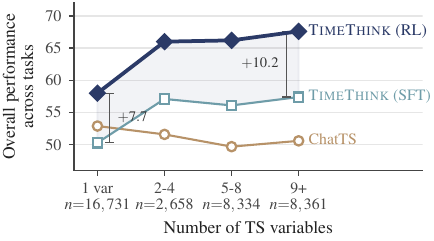}
\nolinenumbers
\caption{Overall performance as the number of TS variables in the prompt grows.}
\label{fig:clustering_sensitivity}
\end{figure}
\noindent\textbf{{Length generalization.}}
To test if \oursRL's gains transfer beyond training-length, we evaluate on unseen range $[769, 1536]$ timesteps and compare overall performance against \oursSFT{} and ChatTS. As shown in \Cref{fig:length-ood-by-category}, \oursRL{} consistently outperforms both baselines across all categories, with especially large gains on compositional tasks, while \oursSFT{} improves over ChatTS on Comp-D but underperforms on atomic tasks and Comp-R.

\noindent\textbf{{Sensitivity analysis w.r.t.\ the number of timeseries.}} We examine how performance changes with the number of timeseries variables in the prompt, splitting the test set into four groups: 1 variable (UTS), 2--4, 5--8, and $\geq 9$ variables. \Cref{fig:clustering_sensitivity} reports overall performance for the three native timeseries variants in each group. \oursRL achieves the highest accuracy across all four groups, and its improvement over \oursSFT{} grows from approximately 8~pp on UTS examples to 10~pp for prompts with $\geq 5$ variables. This suggests RL is particularly effective for multivariate reasoning, where the model must combine evidence across several timeseries.

\noindent\textbf{{Data scaling: SFT vs RL.}} A natural alternative to RL post-training is additional SFT on the same training distribution. We compare the two regimes in \Cref{tab:scaling-axes-story} on SenTSR-Bench~\citep{Heetal2026} and MCQ2~\citep{mcq2} benchmarks. Starting from ChatTS~\citep{xie2024chatts}, adding 110K additional SFT samples improves SenTSR-Bench mean accuracy by only 2.8~pp and \emph{regresses} MCQ2 accuracy by 8.0~pp, consistent with SFT's tendency to reinforce the dominant answer distribution. In contrast, starting from \oursSFT, adding 10K RL samples yields larger gains: $+8.1$~pp on SenTSR-Bench mean accuracy and $+5.7$~pp on MCQ2. The RL recipe thus requires an $11\times$ smaller data budget and only outcome-verifiable supervision rather than reference responses.

\noindent\textbf{{Reward design.}} \Cref{app:reward_design} reports two further case studies: proximity rewards beat exact-match rewards on numerical tasks ($+18.4$ vs.\ $+8.4$~pp), and rewarding composites alone is not enough, since an atomic primitive that is only indirectly rewarded degrades on OOD composites built on it.
\begin{table}[t]
\centering
\caption{Data scaling comparison in SFT vs.\ RL axis. \textsc{What}, \textsc{How} and \textsc{Fix} are the SenTSR-Bench splits.}
\label{tab:scaling-axes-story}
\begin{adjustbox}{max width=0.95\columnwidth,center}
\begin{tabular}{@{}l ccc c c@{}}
\toprule
\multirow{2}{*}{Model} & \multicolumn{4}{c}{\textbf{SenTSR-Bench}} & \multirow{2}{*}{\textbf{MCQ2}} \\
\cmidrule(lr){2-5}
& \textsc{What} & \textsc{How} & \textsc{Fix} & Mean & \\
\midrule
ChatTS                         & 0.236 & 0.218 & \underline{0.273} & 0.242 & 0.420 \\
ChatTS + 110K SFT              & \underline{0.309} & 0.236 & 0.264 & \underline{0.270} & 0.340 \\
\rowcolor{black!4}
$\Delta$ from scaling 110K SFT & \gain{+0.073} & \gain{+0.018} & \loss{-0.009} & \gain{+0.028} & \loss{-0.080} \\
\midrule
\oursSFT{}                     & 0.264 & \underline{0.255} & \underline{0.273} & 0.264 & \underline{0.433} \\
\ours (10K RL)             & \textbf{0.445} & \textbf{0.264} & \textbf{0.327} & \textbf{0.345} & \textbf{0.490} \\
\rowcolor{black!4}
$\Delta$ from scaling 10K RL   & \gain{+0.181} & \gain{+0.009} & \gain{+0.054} & \gain{+0.081} & \gain{+0.057} \\
\bottomrule
\end{tabular}
\end{adjustbox}
\end{table}
\section{Conclusion}
\ours{} grounds timeseries language reasoning in domain-invariant atomic primitives (e.g., trend, seasonality) with deterministic ground truth, which removes costly annotation and makes reinforcement learning with verifiable rewards tractable. The resulting model composes primitives autonomously, outperforming text-based and native timeseries models on both synthetic and real-world benchmarks, and a curriculum over sequence length carries these gains to longer horizons. Primitive-based reasoning is a promising route for high-stakes domains like healthcare and finance.
\section{Limitations}
Given the scope of this work, it necessarily entails certain limitations. \ours{} is trained on a fixed set of synthetic atomic primitives, which may not exhaustively capture all temporal phenomena found in real-world timeseries. Relying on synthetic data is a necessity given the lack of real-world datasets for compositional reasoning with precise ground truth; however, should such resources become available, \ours{} could benefit from training on mixed datasets. As the primary scope of this work is to study compositional reasoning, we do not focus on injecting exhaustive knowledge or various real-world sensor domains. We leave this as a direction for future work.

\section{Acknowledgments}
The authors acknowledge support for this research from \textbf{Evergreen: A Generative AI and Behavioral Sensing Digital Ecosystem to Promote Student Wellness and Flourishing.} This work is made possible through philanthropic gifts
to Dartmouth College dedicated to advancing AI-supported
well-being and flourishing of college students.

\bibliography{custom}

\clearpage
\appendix

\section{Additional Results}

\subsection{Comparison with a Coding Agent}
We also compare with a code-interpreter agent: a 32B model given the raw values of each series and a Python interpreter allowing it to compute statistics exactly. As \Cref{tab:code-agent} shows, the agent performs well on synthetic questions, outperforming \ours on two splits. This is expected because our synthetic questions are defined by simple programs over the raw values, making code execution close to an oracle. Thus, the agent has an advantage unavailable to \ours and is not a like-for-like baseline; we therefore exclude it from the main results. On real-world data, the trend reverses: the agent scores 25.8 on SenTSR-Bench compared with 35.5 for \ours, as these questions require reasoning beyond direct computation over the observed values.

\begin{table}[!ht]
\centering
\caption{Comparison with a code-interpreter agent: Qwen2.5-32B-Instruct given the raw values as text plus a Python interpreter. \ours{} reads the series through its encoder, without tools.}
\label{tab:code-agent}
\resizebox{\columnwidth}{!}{
\begin{tabular}{@{}lcccccc@{}}
\toprule
& \multicolumn{4}{c}{Synthetic} & \multicolumn{2}{c}{Real-world} \\
\cmidrule(lr){2-5}\cmidrule(lr){6-7}
Model & Atomic & Comp-D & Comp-R & OOD & SenTSR-Bench & MCQ2 \\
\midrule
Code agent & \textbf{83.7} & 66.8 & 68.7 & \textbf{72.3} & 25.8 & 42.8 \\
\oursRL & 72.8 & \textbf{73.1} & \textbf{73.4} & 62.9 & \textbf{35.5} & \textbf{45.0} \\
\bottomrule
\end{tabular}
}
\end{table}

\subsection{Additional SFT vs.\ RL}
Starting from the same \oursSFT checkpoint, we compare additional SFT and RL using 10K additional reward-only composite training samples. Since reward-only composite questions do not have gold CoT annotations, we use a rejection-fine-tuning procedure. Specifically, for each prompt, we sample 32 responses from the \oursSFT checkpoint and retain the responses that receive a maximum reward under the same verifiable reward used for GRPO. The resulting data are then used for additional SFT fine-tuning, providing a controlled comparison with RL. As shown in \Cref{tab:additional_sft_rl}, RL achieves the best performance on most metrics, suggesting that RL provides benefits beyond additional fine-tuning on reward-filtered responses.

\begin{table}[!ht]
\centering
\caption{Additional SFT vs.\ RL from the same \oursSFT{} checkpoint, both using 10K additional training samples.}
\label{tab:additional_sft_rl}
\resizebox{\columnwidth}{!}{
\begin{tabular}{@{}lcccccc@{}}
\toprule
Model & Atomic & Comp-D & Comp-R & OOD & SenTSR-Bench & MCQ2 \\
\midrule
\oursSFT & 0.621 & 0.693 & 0.528 & 0.533 & 0.264 & 0.432 \\
+ 10K SFT & 0.611 & 0.681 & 0.567 & 0.529 & \textbf{0.362} & 0.458 \\
+ 10K RL & \textbf{0.648} & \textbf{0.711} & \textbf{0.597} & \textbf{0.564} & 0.345 & \textbf{0.490} \\
\bottomrule
\end{tabular}
}
\end{table}

\subsection{Clustering Task}
Clustering includes two synchronization criteria: \emph{trend}, based on matching or opposite trend-segment patterns, and \emph{local fluctuation}, based on events near a target position. Both criteria appear during SFT phase. \Cref{tab:clustering} shows that \oursSFT already learns this task well, improving average F1 over ChatTS by 25.5 points. RLVR adds only a small gain of 1.1 points. This is consistent with our training design: when a task family is well covered by supervised reasoning traces, most of the improvement comes from SFT, while RLVR has a smaller effect.

\begin{table}[!ht]
\centering
\caption{Clustering by synchronization criterion (F1 metric).}
\label{tab:clustering}
\begin{tabular}{lccc}
\toprule
\multirow{2}{*}{Model} & {Trend} & {Local} & {Avg} \\
      & \scriptsize(\#500) & \scriptsize(\#193) & \scriptsize(\#693) \\
\midrule
ChatTS    & 46.2 & 23.0 & 39.7 \\
\oursSFT  & 70.0 & 52.8 & 65.2 \\
\oursRL   & \textbf{71.3} & \textbf{53.3} & \textbf{66.3} \\
\bottomrule
\end{tabular}
\end{table}

\subsection{Numerical Task}
During RL optimization, numerical tasks (position and magnitude) are scored with a proximity reward rather than exact match, so predictions receive partial credit when they are close to the ground truth. \Cref{tab:numeric-len256} reports relative accuracy for each task. \oursRL is the strongest model in every category. It outperforms the best text-only baseline by 23.1 points on position, 17.1 points on count, and 5.9 points on magnitude. Within the same TS-native models, RLVR also gives consistent gains over \oursSFT, improving position, count, and magnitude by 14.9, 17.2, and 10.7 points, respectively. We hypothesize that CoT supervision encourages step-wise but verbose responses, which are more error-prone on numerical tasks and RL corrects this.

\begin{table}[!ht]
\centering
\caption{Numeric-task relative accuracy; $\Delta_{\mathrm{RLVR}}$ is the RL gain over SFT.}
\label{tab:numeric-len256}
\begin{adjustbox}{max width=\columnwidth,center}
\begin{tabular}{@{}lcccc@{}}
\toprule
\multirow{2}{*}{\textbf{Model}} & \textbf{Position} & \textbf{Count} & \textbf{Magnitude} & \textbf{Avg} \\
 & \scriptsize($\#254$) & \scriptsize($\#379$) & \scriptsize($\#1898$) & \scriptsize($\#2531$) \\
\midrule
Qwen2.5-7B & 55.5 & 45.7 & 50.0 & 49.9 \\
Mistral-7B & 54.0 & 34.9 & 39.8 & 40.5 \\
Time-R1    & 53.0 & 38.8 & 44.6 & 44.6 \\
Time-MQA   & 48.3 & 25.5 & 45.0 & 42.4 \\
\midrule
ChatTS           & 65.6 & 62.2 & 52.7 & 55.4 \\
\oursSFT         & 63.8 & 45.6 & 45.2 & 47.2 \\
\textbf{\oursRL} & \textbf{78.6} & \textbf{62.8} & \textbf{55.9} & \textbf{59.2} \\
\rowcolor{black!4}
$\Delta_{\mathrm{RLVR}}$ & \gain{+14.9} & \gain{+17.2} & \gain{+10.7} & \gain{+12.1} \\
\bottomrule
\end{tabular}
\end{adjustbox}
\end{table}

\subsection{Descriptive Task}
We present the cross-architecture comparison in terms of description performance scored by LLM in \Cref{tab:cross-arch-description}. It demonstrates that \oursRL exhibits superiority in describing the timeseries data.

\begin{table}[!ht]
\centering
\caption{Free-form descriptive task, judge score.}
\label{tab:cross-arch-description}
\begin{tabular}{@{}l c@{}}
\toprule
Model & Description score \\
\midrule
Qwen2.5-7B-Inst                        & 0.456 \\
Mistral-7B-Inst-v0.3                   & 0.464 \\
Time-MQA                               & 0.401 \\
Time-R1                                & 0.414 \\
TimeOmni-1                             & 0.476 \\
\midrule
ChatTS                                 & 0.542 \\
\oursSFT                               & \underline{0.587} \\
\textbf{\oursRL}                     & \textbf{0.708} \\
\bottomrule
\end{tabular}
\end{table}

\subsection{Full Results on SenTSR-Bench}
We present the complete results of SenTSR-Bench in \Cref{tab:sentsr-full} which shows \oursRL, despite being trained only on synthetic data, almost matches the performance of TimeOmni-1 in average performance.

\begin{table}[!ht]
\centering
\caption{Full SenTSR-Bench accuracy. Columns are the \textsc{What Happened}, \textsc{How Happened} and \textsc{Suggested Fix} splits.}
\label{tab:sentsr-full}
\begin{adjustbox}{max width=\columnwidth,center}
\begin{tabular}{@{}l ccc c@{}}
\toprule
Model & \textsc{What} & \textsc{How} & \textsc{Fix} & Mean \\
\midrule
Qwen2.5-7B-Inst                       & 0.218              & 0.200              & 0.282              & 0.233              \\
Mistral-7B-Inst-v0.3                  & 0.273              & \underline{0.300}  & 0.245              & 0.273              \\
Time-MQA                              & 0.255              & 0.245              & 0.227              & 0.242              \\
Time-R1                               & 0.218              & 0.218              & 0.273              & 0.236              \\
TimeOmni-1                            & \underline{0.409}  & \textbf{0.318}     & \textbf{0.364}     & \textbf{0.364}     \\
\midrule
ChatTS                                & 0.236              & 0.218              & 0.273              & 0.242              \\
\oursSFT                              & 0.264              & 0.255              & 0.273              & 0.264              \\
\textbf{\oursRL}                      & \textbf{0.445}     & 0.291              & \underline{0.327}  & \underline{0.355}  \\
\bottomrule
\end{tabular}
\end{adjustbox}
\end{table}

\subsection{Full Results on MCQ2}
We present the complete results of MCQ2 benchmark in \Cref{tab:mcq2-full} which shows \oursRL{} outperforms all baselines except TimeOmni-1.

\begin{table}[!ht]
\centering
\caption{Full MCQ2 evaluation.}
\label{tab:mcq2-full}
\begin{tabular}{@{}l c@{}}
\toprule
Model & Accuracy \\
\midrule
Qwen2.5-7B-Inst & 0.343 \\
Mistral-7B-Inst-v0.3 & 0.270 \\
Time-MQA & 0.366 \\
Time-R1 & 0.305 \\
TimeOmni-1 & \textbf{0.538} \\
\midrule
ChatTS & 0.420 \\
ChatTS+110K SFT & 0.340 \\
\oursSFT & 0.432 \\
\ours{} (10K RL) & \underline{0.490} \\
% \oursRL & 0.450 \\
\bottomrule
\end{tabular}
\end{table}

\subsection{Reward Design Case Studies}
\label{app:reward_design}
\noindent\textbf{Proximity-based vs.\ exact-match reward.} Two periodicity tasks in our benchmark enable a comparison. Both require the model to output a single scalar derived from a timeseries: either the period of the series or the number of complete cycles within a given window. We assign a proximity reward to the former and an exact-match reward to the latter. Empirically, RL training yields a gain of only $+8.4$~pp on the exact-match task, compared to $+18.4$~pp on the proximity-rewarded counterpart, suggesting the advantage of proximity-based rewards for numerical tasks.

\noindent\textbf{Is rewarding composite QAs enough?} We study this in a setting where the atomic primitive receives only indirect supervision. \oursSFT{} supervises trend-segment enumeration of the form \texttt{[('increase', start, end), ('decrease', start, end), \ldots]}. During \oursRL{}, we restrict reward on this atom to count-based QAs over the segment list. The drift is sharp on OOD composites that require precise trend-segment positions: \oursRL{} underperforms \oursSFT{} by $18.9$~pp on \textsc{cross trend convergence} (comparing trend agreement across temporal halves) and by $13.4$~pp on \textsc{trend reversal} (detecting whether the dominant trend reverses at the largest change point). Both compositions decompose through the same primitive whose positional structure was never directly rewarded under our count-only RL signal. Atomic primitives should therefore be rewarded directly; otherwise the policy degrades on them, and the OOD composites built on them suffer most.

\clearpage
\onecolumn
\section{Qualitative Example}
\begin{figure}[h]
    \centering
    \includegraphics[width=\linewidth]{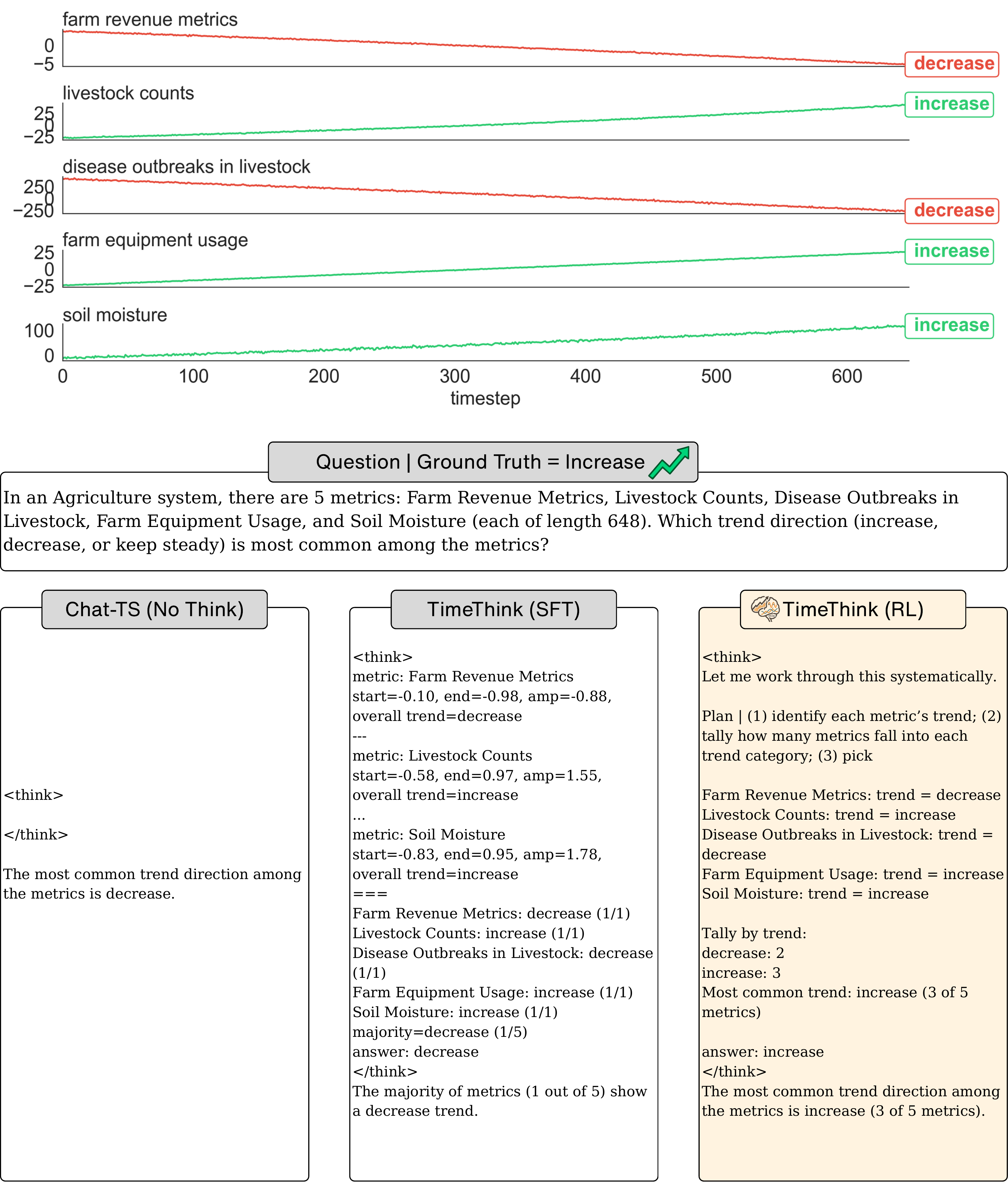}
    \caption{Comparison of responses from ChatTS, \oursSFT{} and \oursRL{} on the same question.}
    \label{fig:qualitative_results}
\end{figure}

\Cref{fig:qualitative_results} compares the responses of ChatTS, \oursSFT{} and \oursRL{} on the same question.

\twocolumn

\section{LLM-as-Judge Prompt Templates}
\label{app:judge-prompts}

All free-form and structured-verdict evaluations are graded by
\texttt{Qwen2.5-72B-Instruct-GPTQ-Int4} served via vLLM with greedy
decoding. The judge is invoked through one of four prompt templates
selected by the task family of each evaluation sample. We reproduce
each template verbatim below; placeholder fields (e.g.\ \texttt{\{question\}},
\texttt{\{pred\}}) are filled with sample-specific content at inference
time. Outputs are parsed as JSON.

% --------------------------------------------------------------------
\paragraph{Free-form judge.}
Used for the descriptive task and for the TSEvol's inductive tasks.
Returns a single graded score for explanation quality.

\begin{tcolorbox}[colback=gray!4, colframe=gray!50, boxrule=0.4pt,
  arc=2pt, left=6pt, right=6pt, top=4pt, bottom=4pt, breakable,
  fontupper=\small\ttfamily]
You are an expert time-series analyst evaluator.\\
Your task is to rate the quality of a generated explanation compared to a ground truth explanation.\\[2pt]
Question: \{question\}\\
Ground Truth: \{gt\}\\
Prediction: \{pred\}\\[2pt]
Critique the prediction based on physical significance and accuracy.\\
Finally, provide a score from 0.0 to 1.0.\\[2pt]
Output ONLY the JSON object. Do not include any introductory text, markdown headers, or explanations outside of the JSON block:\\
\{\\
\hspace*{1.5em}"reasoning": "your critique here",\\
\hspace*{1.5em}"score": 0.85\\
\}
\end{tcolorbox}

% --------------------------------------------------------------------
\paragraph{Multiple-choice judge.}
Used for MCQ2, SenTSR-Bench.
The judge first identifies which option the model committed to, then sets a binary
\texttt{verdict\_match} and a separate explanation-quality score.

\begin{tcolorbox}[colback=gray!4, colframe=gray!50, boxrule=0.4pt,
  arc=2pt, left=6pt, right=6pt, top=4pt, bottom=4pt, breakable,
  fontupper=\small\ttfamily]
You are grading a multiple-choice answer.\\[2pt]
Question:\\
\{question\}\\[2pt]
Options:\\
\{options\_block\}\\[2pt]
Correct answer: \{correct\_letter\}) \{correct\_option\}\\[2pt]
Model response:\\
\{pred\}\\[2pt]
Task:\\
1. Identify which option (A/B/C/D) the model ultimately picked. Acceptable forms:\\
\hspace*{1.5em}-- Starts with the letter (e.g.\ "A) ...", "Answer: B", "The answer is C").\\
\hspace*{1.5em}-- Paraphrases or quotes the option text without the letter — match by content.\\
\hspace*{1.5em}-- Rambles or revises — use the final committed choice (the answer it ends on).\\
\hspace*{1.5em}-- Does NOT commit to any option $\rightarrow$ picked\_letter $=$ null.\\
2. Set verdict\_match $=$ 1.0 iff picked\_letter $==$ correct\_letter, else 0.0. If picked\_letter is null, verdict\_match $=$ 0.0.\\
3. Rate reasoning quality in score (0.0--1.0) INDEPENDENT of correctness — how clear, grounded in the time series, and well-structured the justification is.\\[2pt]
Output ONLY the JSON object. No preface, no markdown:\\
\{\\
\hspace*{1.5em}"picked\_letter": "A",\\
\hspace*{1.5em}"verdict\_match": 1.0,\\
\hspace*{1.5em}"reasoning": "brief justification for the grading",\\
\hspace*{1.5em}"score": 0.85\\
\}
\end{tcolorbox}

% --------------------------------------------------------------------
\paragraph{Structured-verdict judge.}
Used for our compositional task families (segment-family, list-and-set
tasks, and numerical-verdict compositions). Returns four signals: a
binary \texttt{verdict\_match}, a graded \texttt{correctness\_score}
(Kendall's tau intuition for ordered lists, Jaccard intuition for sets),
an overall explanation \texttt{score}, and an \texttt{extracted\_number}
for numeric-verdict tasks (e.g., proportions stated as ``3 of 7'' are
auto-converted to $0.43$).

\begin{tcolorbox}[colback=gray!4, colframe=gray!50, boxrule=0.4pt,
  arc=2pt, left=6pt, right=6pt, top=4pt, bottom=4pt, breakable,
  fontupper=\small\ttfamily]
You are an expert time-series analyst evaluator.\\
Your task is to evaluate a model's response to a structured time-series question.\\[2pt]
Question: \{question\}\\
Expected Answer: \{expected\_verdict\}\\
Ground Truth Response: \{gt\}\\
Model Response: \{pred\}\\[2pt]
Evaluate FOUR things:\\[2pt]
1. \textbf{verdict\_match}: Does the model's response convey the same answer/verdict as the expected answer? Use 1.0 for correct, 0.0 for incorrect. Be lenient with formatting -- ``yes'', ``Yes.'', and ``Yes, because\ldots'' all match an expected verdict of ``yes''. For numeric verdicts, allow small rounding differences.\\[2pt]
2. \textbf{correctness\_score}: A graded measure of ANSWER CORRECTNESS only -- ignore explanation quality, depth, or verbosity. Range 0.0--1.0:\\
\hspace*{1.5em}-- 1.0 $=$ fully matches the expected answer (any acceptable phrasing)\\
\hspace*{1.5em}-- 0.7--0.9 $=$ mostly correct with a small/specific error (one swap in a 6-list, a partially-overlapping set)\\
\hspace*{1.5em}-- 0.4--0.6 $=$ roughly half right (half a list correct, right order of magnitude but off)\\
\hspace*{1.5em}-- 0.1--0.3 $=$ mostly wrong but some element is recognizable\\
\hspace*{1.5em}-- 0.0 $=$ wrong, or no answer\\
For ordered lists, use Kendall's tau intuition (more swaps $\rightarrow$ lower score). For sets, use Jaccard intuition ($|$intersection$|/|$union$|$).\\[2pt]
3. \textbf{score}: Overall explanation quality (reasoning, accuracy, completeness) from 0.0 to 1.0. Diagnostic; can differ from correctness\_score (a terse-but-correct answer can have low score and high correctness\_score).\\[2pt]
4. \textbf{extracted\_number}: If the Expected Answer is numeric, identify the final numeric verdict the model is asserting -- the number it CLAIMS as its answer, not an intermediate calculation. For ratios in words (``3 of 7''), compute to 2 decimal places ($\approx 0.43$). Output as a JSON number; null if no clear numeric verdict.\\[2pt]
Output ONLY the JSON object:\\
\{\\
\hspace*{1.5em}"reasoning": "your critique here",\\
\hspace*{1.5em}"verdict\_match": 1.0,\\
\hspace*{1.5em}"correctness\_score": 0.85,\\
\hspace*{1.5em}"score": 0.85,\\
\hspace*{1.5em}"extracted\_number": 0.52\\
\}
\end{tcolorbox}

% --------------------------------------------------------------------
\paragraph{Description-extraction judge.}
For the descriptive task we additionally extract a structured
representation of the model's free-form description (trend, seasonality,
noise, local events, physical explanation), allowing per-axis precision
and recall against the ground-truth structured fields. This complements
the free-form judge above; the per-axis F1 contributes to the
\texttt{description\_overall} column reported in our descriptive-task table.

\begin{tcolorbox}[colback=gray!4, colframe=gray!50, boxrule=0.4pt,
  arc=2pt, left=6pt, right=6pt, top=4pt, bottom=4pt, breakable,
  fontupper=\small\ttfamily]
Extract structured time series description data from the following text.\\[2pt]
IMPORTANT: Read the ENTIRE response holistically before extracting. Do NOT do raw keyword extraction. Instead, logically reason about the content:\\
\hspace*{1.5em}-- If ``spike'' is mentioned in one sentence and ``followed by increase'' in another, understand them as related context.\\
\hspace*{1.5em}-- Synthesize scattered descriptions into coherent structured fields.\\[2pt]
Only fill in perspectives that are actually discussed in the text. Leave others as null.\\[2pt]
For \textbf{trend}: identify the overall type (increasing/decreasing/steady/multiple), start value, end value, amplitude, and any segments.\\
For \textbf{seasonality/periodicity}: identify whether periodicity exists, its type, period length, and amplitude.\\
For \textbf{noise}: identify the noise level category (smooth or noisy) and the noise strength value (ratio of signal amplitude).\\
For \textbf{local characteristics}: identify each local event with its type (spike, dip, shake, sudden increase/decrease, etc.), approximate position, amplitude, and any explanation of physical meaning specific to that event.\\
For \textbf{physical explanation}: extract any text that explains the physical meaning or significance of the time series characteristics.\\[2pt]
Return a valid JSON object matching the output schema.
\end{tcolorbox}
\section{QA Types}

\subsection{Atomic QA}
\label{app:atomic}
Atomic QA exercises a single primitive operator $f \in \mathcal{F}_{\mathrm{atom}}$ extracting one fact directly from the timeseries. We instantiate $|\mathcal{F}_{\mathrm{atom}}| = 9$ primitives across 14 single-metric question variants.

\begin{itemize}
    \item \textsc{Mean}: \emph{``What mean value does {\metric} take?''}
    \item \textsc{Mean}: \emph{``What is the average value of {\metric} from position $a$ to $b$?''}
    \item \textsc{Mean} (per chunk): \emph{``What are the means of consecutive 16-point chunks of {\metric} from index $a$ to $b$?''}
    \item \textsc{Std}: \emph{``What is the spread of {\metric} measured by standard deviation?''}
    \item \textsc{Std}: \emph{``What is the standard deviation of {\metric} over $[a, b]$?''}
    \item \textsc{Percentile}: \emph{``Report the 50\textsuperscript{th} percentile value for {\metric}.''}
    \item \textsc{ExtVal}: \emph{``What is the largest value of {\metric}?''}
    \item \textsc{ExtVal}: \emph{``What is the minimum value of {\metric}?''}
    \item \textsc{ExtPos}: \emph{``Where is the highest value in {\metric}?''}
    \item \textsc{ExtPos}: \emph{``At which index does {\metric} attain its minimum?''}
    \item \textsc{EventEnum}: \emph{``Enumerate the local events in {\metric}, including type, position, amplitude.''}
    \item \textsc{SegEnum}: \emph{``Describe the segments of the trend in {\metric} over time.''}
    \item \textsc{Period}: \emph{``Describe the periodic behavior of {\metric}.''}
    \item \textsc{TrendClassify}: \emph{``What is the overall trend direction of {\metric} (increase, decrease, or keep
steady)?''}
  \end{itemize}

\subsection{Sub-skills}
\label{app:subskills}
Sub-skills $\mathcal{F}_{\mathrm{op}}$ act on the outputs of atoms (or earlier sub-skills), not on the raw series. We instantiate $|\mathcal{F}_{\mathrm{op}}| = 9$.
\begin{itemize}
    \item \textsc{Count}: cardinality of a (possibly filtered) list.
    \item \textsc{Sum}: sum over a list of weighted scalars (used for total durations, mode counts, weighted means).
    \item \textsc{Filter}: subset of a list matching a predicate.
    \item \textsc{Locate}: find the structure (segment / chunk / half / quarter / interval) containing a given index, or the bin a derived value falls in.
    \item \textsc{Compare}: binary inequality between two scalars or labels.
    \item \textsc{Threshold}: $x > c$ for a constant $c$.
    \item \textsc{Argmax}/\textsc{Argmin}: select the element with the largest/smallest key.
    \item \textsc{Rank}: total order over a list by a key.
    \item \textsc{Arith}: scalar arithmetic over reals: $+, -, \times, \div, |\cdot|$.
\end{itemize}

\subsection{Comp-D compositions (36)}
\label{app:easy}
\begin{itemize}
    \item \textsc{Mean}/\textsc{Std} on two metrics + \textsc{Compare}: \emph{``Is {\metricB}'s mean higher than {\metricA}'s?''}
    \item \textsc{Mean}/\textsc{Std} on each metric + \textsc{Argmax}: \emph{``Identify the metric with the largest standard deviation.''}
    \item \textsc{TrendClassify} on each metric + \textsc{Filter}: \emph{``Which metrics show an increase trend?''}
    \item \textsc{Mean}/\textsc{TrendClassify} on each metric + \textsc{Count}: \emph{``Count the metrics whose mean exceeds $\tau$.''}
    \item \textsc{TrendClassify} on two metrics + \textsc{Compare}: \emph{``Is the overall trend of {\metricA} the same as {\metricB}?''}
    \item $\textsc{EventEnum} \to \textsc{Locate}$ (within tolerance) $\to \textsc{Threshold}({>}0)$. \emph{``Does {\metric} have a noticeable event around point $p$ within tolerance $\tau$?''}
    \item $\textsc{Period}$ (and \textsc{Count} for cycle count). \emph{``Compute the estimated period for {\metric}.''}
    \item $\textsc{SegEnum} \to \textsc{Argmax}$ (largest level shift) and/or $\textsc{Count}$. \emph{``How many times does {\metric} change its dominant trend?''}
    \item $\textsc{Mean}/\textsc{Std}/\textsc{Percentile}$ on sub-intervals + $\textsc{Arith}$/$\textsc{Compare}$. \emph{``How does the std of {\metric} change from the first half to the second half?''}
    \item multi-condition compound: $\textsc{Mean}{\times}2 + \textsc{Arith} + \textsc{Threshold}$. \emph{``If a `significant upward shift' is when (mean(Q3)-mean(Q1))/range > 15\%, does {\metric} qualify?''}
    \item $\textsc{SegEnum} \to \textsc{Filter}$ (intersect $[a,b]$) $\to \textsc{Sum}$ per type $\to \textsc{Argmax}$. \emph{``Between points $a$ and $b$ in {\metric}, which trend behavior dominates?''}
    \item $\textsc{EventEnum} \to \{\textsc{Filter}, \textsc{Count}, \textsc{Argmax}\}$. \emph{``What is the shortest local event in {\metric} in terms of duration?''}
    \item $\textsc{SegEnum} \to \{\textsc{Filter}, \textsc{Count}, \textsc{Argmax}\}$. \emph{``How many trend segments show an increase in {\metric}?''}
    \item $\textsc{SegEnum} \to$ adjacent-pair extraction $\to \{\textsc{Filter}/\textsc{Count}, \textsc{Argmax}\}$. \emph{``Which adjacent segment pair pattern is most common in {\metric}?''}
    \item $\textsc{EventEnum}, \textsc{SegEnum} \to \textsc{Filter}/\textsc{Locate} \to \{\textsc{Count}, \textsc{Argmax}\}$. \emph{``Count the local events in {\metric} that fall within decrease segments.''}
    \item $\textsc{EventEnum} \to \textsc{Locate}$ (first/last/gap). \emph{``Where does the last local event appear in {\metric}?''}
    \item $\textsc{SegEnum} \to \textsc{Filter}$ per type $\to \textsc{Sum}$ + $\textsc{Compare}$. \emph{``In {\metric}, which has a longer total duration: increase or decrease?''}
    \item $\textsc{TrendClassify}$ on two metrics over aligned windows + similarity check. \emph{``Evaluate the trend similarity between {\metricA} and {\metricB}; transitions within 2 timesteps are aligned.''}
    \item $\textsc{TrendClassify}$ on two metrics + opposite-trend predicate. \emph{``Determine whether {\metricA} and {\metricB} exhibit opposite trend behavior.''}
    \item $\textsc{TrendClassify}$/$\textsc{EventEnum}$ on each metric + similarity to anchor + $\textsc{Filter}$. \emph{``Identify metrics whose local behavior near point $p$ resembles {\metricAnchor}.''}
    \item $\textsc{TrendClassify}$ on each metric + opposite-anchor $\textsc{Filter}$. \emph{``Find metrics that move opposite to {\metricAnchor}.''}
    \item atom on each metric + $\{\textsc{Filter}, \textsc{Argmax}\}$. \emph{``Which metric has the highest standard deviation?''}
    \item atom on two metrics + $\textsc{Arith}$ (ratio/difference) + $\textsc{Threshold}$. \emph{``An `instability gap' occurs when std({\metricA}) is at least $r\times$ std({\metricB}); does it occur?''}
    \item anticorrelation predicate $\wedge$ noise-strength threshold. \emph{``An `anticorrelated high-noise state' is when {\metricA} and {\metricB} are anti-correlated AND noise of {\metricB} exceeds $\tau$; does it hold?''}
    \item $\textsc{TrendClassify}$ on {\metricA} + conditional $\textsc{TrendClassify}$ on {\metricB} over the same windows. \emph{``When {\metricA} is increasing, what does {\metricB}'s trend look like?''}
    \item $\textsc{Mean}{\times}2 \to \textsc{Compare}$. \emph{``Does the first-half mean of {\metric} exceed the second-half mean?''}
    \item $\textsc{Std}{\times}2 \to \textsc{Compare}$. \emph{``Is the std of the second half of {\metric} higher than the first?''}
    \item $\textsc{EventEnum}, \textsc{SegEnum} \to \textsc{Locate}/\textsc{Filter} \to \textsc{Threshold}({>}0)$. \emph{``Does any local event of {\metric} fall in an increase segment?''}
    \item $\textsc{ExtPos}, \textsc{SegEnum} \to \textsc{Locate} \to \textsc{Compare}$ on type. \emph{``Does the global max of {\metric} fall within a decrease segment?''}
    \item $\textsc{EventEnum}, \textsc{Std} \to \textsc{Argmax} \to \textsc{Arith}({\times}k) \to \textsc{Threshold}$. \emph{``Is the seasonal amplitude of {\metric} greater than $k\sigma$?''}
    \item $\textsc{EventEnum} \to \textsc{Filter} \to \textsc{Count}$. \emph{``How many decrease events are there in {\metric}?''}
    \item $\textsc{SegEnum} \to \textsc{Argmax}$ (duration) $\to$ extract type. \emph{``Is the longest segment in {\metric} an increase, decrease, or steady segment?''}
    \item $\textsc{SegEnum} \to \textsc{Filter} \to \textsc{Sum} \to \textsc{Arith}({\div}\,\text{total})$. \emph{``What share of {\metric} is decreasing?''}
    \item $\textsc{Std}{\times}2 \to \textsc{Arith}({\div}) \to \textsc{Threshold}$. \emph{``Does {\metricA} have a std at least $r\times$ that of {\metricB}?''}
    \item $\textsc{Mean}{\times}N \to \textsc{Rank}$. \emph{``Rank the metrics by mean from highest to lowest.''}
    \item $\textsc{TrendClassify}{\times}N \to \textsc{Filter}/\textsc{Count}$ per type $\to \textsc{Argmax} \to \textsc{Arith}({\div}\,N)$. \emph{``How dominant is the most common trend? Report the mode share.''}
\end{itemize}

\subsection{Comp-R compositions (32)}
\label{app:hard}
\begin{itemize}
    \item $\textsc{EventEnum} \to \textsc{Count}$. \emph{``How many spikes, dips, or sudden changes are in {\metric}?''}
    \item $\textsc{SegEnum} \to \textsc{Count}$. \emph{``How many distinct trend segments does {\metric} have?''}
    \item $\textsc{SegEnum} \to \textsc{Locate}$ (by index) $\to$ duration. \emph{``How many points are in the segment of {\metric} containing index $p$?''}
    \item $\textsc{SegEnum} \to \textsc{Locate} \to$ type. \emph{``What is happening in {\metric} at index $p$?''}
    \item $\textsc{SegEnum}, \textsc{Mean}{\times}2 \to \textsc{Locate}{\times}2 \to \textsc{Compare}$. \emph{``Is the mean of the segment at index $p$ greater than at $q$?''}
    \item $\textsc{SegEnum} \to \textsc{Argmax}$ (duration). \emph{``What is the duration of the longest trend segment in {\metric}?''}
    \item $\textsc{SegEnum} \to \textsc{Filter}/\textsc{Count}$ per type $\to \textsc{Argmax}$. \emph{``What is the dominant trend type by segment count in {\metric}?''}
    \item $\textsc{EventEnum} \to \textsc{Argmax} \to$ type. \emph{``Find the local event of {\metric} with the largest amplitude; report its type.''}
    \item $\textsc{EventEnum} \to \textsc{Locate}$ (nearest) $\to$ type. \emph{``What is the nearest event type to position $p$ in {\metric}?''}
    \item $\textsc{ExtPos}, \textsc{EventEnum} \to \textsc{Locate}$ (window) $\to \textsc{Threshold}({>}0)$. \emph{``Does a local event of {\metric} occur near its global max or min?''}
    \item $\textsc{ExtPos}{\times}2 \to \textsc{Locate} \to \textsc{Compare}$. \emph{``Are the global max and min of {\metric} in the same half?''}
    \item $\textsc{ExtPos} \to \textsc{Locate}$ (half). \emph{``Does the global max of {\metric} occur before the midpoint?''}
    \item $\textsc{ExtVal}{\times}2 \to \textsc{Arith}({-})$. \emph{``What is the spread of {\metric} (max minus min)?''}
    \item $\textsc{ExtVal}{\times}2, \textsc{Std} \to \textsc{Arith}({-},{\div})$. \emph{``What is the normalized range of {\metric}?''}
    \item $\textsc{EventEnum}, \textsc{ExtVal}{\times}2 \to \textsc{Argmax} \to \textsc{Arith} \to \textsc{Threshold}$. \emph{``Does the peak event amplitude of {\metric} exceed half the value range?''}
    \item $\textsc{Mean}{\times}2 \to \textsc{Arith}(|{-}|)$. \emph{``Compute $|\mathrm{mean}(\text{first half}){-}\mathrm{mean}(\text{second half})|$.''}
    \item $\textsc{Mean}{\times}2$ (thirds) $\to \textsc{Arith}(|{-}|) \to \textsc{Threshold}$. \emph{``Is there a mean shift greater than $\tau$ between the first and last thirds of {\metric}?''}
    \item $\textsc{Mean}{\times}2$ ($Q_1, Q_4$) $\to \textsc{Arith}(|{-}|) \to \textsc{Threshold}({<}\tau)$. \emph{``Is $|\mathrm{mean}(Q_1){-}\mathrm{mean}(Q_4)| < \tau$ for {\metric}?''}
    \item $\textsc{Mean}, \textsc{Mean}$ over chunks $\to \textsc{Filter}$ (within $\tau$) $\to \textsc{Compare}$ to $|L|$. \emph{``Are all 16-point chunk means of {\metric} within $\tau$ of the global mean?''}
    \item $\textsc{Percentile}(50), \textsc{Mean} \to \textsc{Arith}(|{-}|) \to \textsc{Threshold}$. \emph{``Is {\metric}'s median within $\tau$ of its mean?''}
    \item $\textsc{Mean}{\times}2 \to \textsc{Compare}$. \emph{``Is the mean of {\metric} on $[a_1,b_1]$ greater than on $[a_2,b_2]$?''}
    \item $\textsc{Period} \to \textsc{Count}$. \emph{``Count the approximate number of complete cycles in {\metric}.''}
    \item $\textsc{Period}$. \emph{``Estimate the period of {\metric} in time steps.''}
    \item $\textsc{Period} \to \textsc{Threshold}$. \emph{``Does {\metric} contain a periodic component?''}
    \item $\textsc{TrendClassify}{\times}N \to \textsc{Filter}/\textsc{Count}$ per type $\to \textsc{Argmax}$. \emph{``Which trend pattern is most common across the metrics?''}
    \item $\textsc{TrendClassify}{\times}N \to$ distinct count. \emph{``Into how many groups do the metrics cluster by trend?''}
    \item $\textsc{TrendClassify}{\times}N \to \textsc{Filter}/\textsc{Count}$ per type $\to \textsc{Argmax}$. \emph{``What is the dominant trend cluster?''}
    \item $\textsc{EventEnum}{\times}2 \to \textsc{Locate}$ (within $w$) $\to \textsc{Threshold}({>}0)$. \emph{``Do {\metricA} and {\metricB} have temporally close events?''}
    \item $\textsc{Period}{\times}2 \to \textsc{Compare}$. \emph{``Does {\metricA} oscillate faster than {\metricB}?''}
    \item $\textsc{Mean}{\times}N, \textsc{Std}{\times}N \to \textsc{Argmax}{\times}2 \to \textsc{Compare}$ (equal). \emph{``Is the metric with the highest mean also the one with the highest std?''}
    \item $\textsc{TrendClassify}{\times}N, \textsc{Mean}/\textsc{Std}{\times}N \to \textsc{Filter} \to \textsc{Argmax}$. \emph{``Among metrics with an increase trend, which has the largest range?''}
    \item $\textsc{Mean}{\times}2{\times}N \to \textsc{Compare}$ per metric $\to \textsc{Filter}+\textsc{Count}$. \emph{``How many metrics show an upward shift in mean between halves?''}
\end{itemize}

\subsection{OOD compositions (34)}
\label{app:ood}
\begin{itemize}
    \item $\textsc{ExtPos}{\times}2 \to \textsc{Compare}$. \emph{``Is the index of the global max before the index of the global min?''}
    \item $\textsc{Mean}{\times}4, \textsc{ExtPos} \to \textsc{Argmax} \to \textsc{Locate} \to \textsc{Compare}$. \emph{``Does the global max fall in the quarter with the highest mean?''}
    \item $\textsc{Mean}{\times}4 \to \textsc{Rank}$. \emph{``Do the quarter means of {\metric} strictly increase (or decrease)?''}
    \item $\textsc{Mean}$ over chunks $\to \textsc{Argmax}$. \emph{``Find the 16-point chunk with the largest mean and report its index.''}
    \item $\textsc{Mean}$ over chunks, $\textsc{Mean} \to \textsc{Filter} \to \textsc{Count} \to \textsc{Arith}({\div}\,|L|)$. \emph{``What fraction of 16-point chunks have a mean above the global mean?''}
    \item $\textsc{Std}, \textsc{ExtVal}{\times}2 \to \textsc{Arith} \to \textsc{Threshold}$. \emph{``Is the standard deviation greater than $\tfrac{1}{3}$ of the range?''}
    \item $\textsc{Mean}{\times}4 \to \textsc{Arith}(|{-}|){\times}2 \to \textsc{Threshold}{\times}2 \to \wedge$. \emph{``Does {\metric} exhibit symmetric recovery ($Q_1{\approx}Q_4$ and $Q_2{\approx}Q_3$)?''}
    \item $\textsc{Period}, \textsc{Mean}$ per cycle $\to \textsc{TrendClassify}$ on cycle-mean list. \emph{``Do the cycle-level means show an increasing or decreasing trend?''}
    \item $\textsc{EventEnum}, \textsc{SegEnum}, \textsc{Std} \to \textsc{Argmax} \to \textsc{Locate} \to \textsc{Threshold}$. \emph{``Does the largest event amplitude exceed the std of its containing segment?''}
    \item $\textsc{EventEnum}, \textsc{Std} \to \textsc{Argmax} \to \textsc{Arith}({\times}k) \to \textsc{Threshold}$. \emph{``Does the largest event amplitude exceed $k\sigma$?''}
    \item $\textsc{EventEnum}, \textsc{ExtVal}{\times}2 \to \textsc{Argmax} \to \textsc{Arith} \to \textsc{Threshold}$. \emph{``Does the largest event amplitude exceed half the value range?''}
    \item $\textsc{EventEnum}, \textsc{SegEnum} \to \textsc{Filter} \to \textsc{Count}/\textsc{Sum} \to \textsc{Arith}({\div}) \to \textsc{Argmax}/\textsc{Argmin}$. \emph{``Which trend type has the lowest density of local events?''}
    \item $\textsc{EventEnum}, \textsc{SegEnum} \to \textsc{Filter} \to \textsc{Count}, \textsc{Sum} \to \textsc{Arith}({\div})$. \emph{``What is the event density in increase segments?''}
    \item $\textsc{SegEnum}, \textsc{EventEnum} \to \textsc{Argmax} \to \textsc{Filter} \to \textsc{Argmax}$. \emph{``What is the largest event amplitude in the longest trend segment?''}
    \item $\textsc{SegEnum}, \textsc{ExtPos} \to \textsc{Argmax} \to \textsc{Locate}$. \emph{``Does the global max fall inside the longest trend segment?''}
    \item $\textsc{SegEnum} \to \textsc{Sum}$ (weighted) $\to \textsc{Arith}({\div}\,\text{total})$. \emph{``Compute the duration-weighted mean across trend segments.''}
    \item $\textsc{SegEnum} \to \textsc{Argmax} \to \textsc{Filter}$ (same type) $\to \textsc{Sum} \to \textsc{Arith}({\div})$. \emph{``What fraction of {\metric} is covered by the trend type of the longest segment?''}
    \item $\textsc{SegEnum} \to \textsc{Filter} \to \textsc{Mean}$ on union. \emph{``Compute the mean of {\metric} restricted to increase segments.''}
    \item $\textsc{SegEnum} \to \textsc{Filter} \to \textsc{Sum} \to \textsc{Arith}({\div}\,\text{count})$. \emph{``What is the average of segment means for all increase segments?''}
    \item $\textsc{SegEnum}, \textsc{Mean} \to \textsc{Filter} \to \textsc{Threshold} \to \textsc{Count}$. \emph{``Count increase segments whose mean exceeds the global mean.''}
    \item $\textsc{SegEnum} \to \textsc{Locate}$ (first/last) $\to \textsc{Compare}$. \emph{``Is the mean of the first segment greater than the last segment?''}
    \item $\textsc{SegEnum}, \textsc{Mean} \to \textsc{Filter} \to \textsc{Compare}\ (\forall)$. \emph{``Do all increase segments have above-average means?''}
    \item $\textsc{SegEnum} \to \textsc{Argmax}$ (level shift) $\to \textsc{Compare}$ on types. \emph{``Does the dominant trend reverse after the largest change point?''}
    \item $\textsc{SegEnum} \to$ palindrome via pairwise $\textsc{Compare}$. \emph{``Is the trend pattern of {\metric} palindromic?''}
    \item $\textsc{TrendClassify}{\times}N \to$ pairwise $\textsc{Compare} \to \textsc{Count}$. \emph{``How many pairs of metrics share the same overall trend direction?''}
    \item $\textsc{TrendClassify}{\times}N$ over halves $\to \textsc{Count} \to \textsc{Compare}$. \emph{``Do the metrics become more aligned in trend in the second half than the first?''}
    \item $\textsc{ExtPos}{\times}N \to$ pairwise $\textsc{Locate}$ within window. \emph{``Do {\metricA} and {\metricB} have their maxima at similar positions?''}
    \item $\textsc{ExtVal}{\times}2{\times}N \to \textsc{Compare}$ on interval ends. \emph{``Is there any overlap between the value ranges of two metrics?''}
    \item $\textsc{Mean}{\times}2{\times}N \to \textsc{Compare}$ per metric $\to \textsc{Count}$. \emph{``How many metrics show an upward shift in mean between halves?''}
    \item $\textsc{TrendClassify}{\times}N \to \textsc{Filter}/\textsc{Count}\ \exists$. \emph{``Is any metric in its own unique trend cluster?''}
    \item $\textsc{TrendClassify}{\times}3 \to \textsc{Compare}{\times}2 \to \wedge$. \emph{``If A's trend matches B's and B's matches C's, does A's match C's?''}
    \item $\textsc{TrendClassify}{\times}N \to$ pairwise opposite check $\to \textsc{Count}$. \emph{``Count the anti-correlated pairs among the metrics.''}
    \item $\textsc{Mean}{\times}N, \textsc{TrendClassify} \to \textsc{Filter}$ (upper half) $\to \textsc{Filter}/\textsc{Count}$ per type $\to \textsc{Argmax}$. \emph{``Among metrics in the upper-half mean group, identify the dominant trend type.''}
    \item $\textsc{EventEnum}{\times}2 \to \textsc{Locate}$ (precedence within $w$). \emph{``Does an event in {\metricA} precede an event in {\metricB} within $w$ timesteps?''}
\end{itemize}
\section{Compositional Performance Breakdown}
\label{app:comp_breakdown}
\Cref{tab:comp-id-easy,tab:comp-id-hard,tab:comp-ood} give the per-composition breakdown behind \Cref{fig:rl_gain}, for the Comp-D, Comp-R and OOD splits respectively. For each composition we report the score of \oursSFT{} and \oursRL{} on its constituent atoms (atomic gain $\Delta_a$) and on the composition itself (composition gain $\Delta_c$).

\begin{table*}[t]
\centering
\footnotesize
\setlength{\tabcolsep}{6pt}
\caption{Compositional reasoning in Comp-D QAs. Composition gains $\Delta_c$ are small on average because SFT CoT supervision already teaches the decomposition.}
\label{tab:comp-id-easy}
\begin{adjustbox}{max width=\textwidth,center}
\begin{tabular}{@{}l ccc c ccc@{}}
\toprule
\multirow{2}{*}{Task} & \multicolumn{3}{c}{Atomic skills} & & \multicolumn{3}{c}{Composition} \\
\cmidrule(lr){2-4} \cmidrule(lr){6-8}
 & \oursSFT & \oursRL & $\Delta_a$ & & \oursSFT & \oursRL & $\Delta_c$ \\
\midrule
Events within trend type & 0.48 & 0.60 & +0.12 & & 0.20 & 0.60 & +0.40 \\
Max within trend type & 0.56 & 0.61 & +0.05 & & 0.55 & 0.72 & +0.17 \\
Type of longest segment & 0.34 & 0.40 & +0.06 & & 0.64 & 0.80 & +0.16 \\
Volatility change ($\times$2 windows) & 0.56 & 0.68 & +0.13 & & 0.61 & 0.76 & +0.15 \\
First-vs-second-half mean & 0.70 & 0.83 & +0.13 & & 0.74 & 0.87 & +0.12 \\
Amplitude vs std ratio & 0.65 & 0.71 & +0.06 & & 0.64 & 0.74 & +0.10 \\
Count events by type & 0.62 & 0.79 & +0.18 & & 0.50 & 0.60 & +0.10 \\
Trend-type duration fraction & 0.34 & 0.40 & +0.06 & & 0.79 & 0.86 & +0.07 \\
Numeric statistic (poly) & 0.68 & 0.78 & +0.10 & & 0.46 & 0.53 & +0.07 \\
Enumerate matching metrics & 0.57 & 0.74 & +0.17 & & 0.65 & 0.72 & +0.07 \\
Enumerate local events & 0.62 & 0.79 & +0.18 & & 0.42 & 0.47 & +0.04 \\
Anti-correlation predicate & 0.68 & 0.92 & +0.23 & & 0.81 & 0.85 & +0.04 \\
Segment-trend predicate & 0.34 & 0.40 & +0.06 & & 0.81 & 0.84 & +0.03 \\
Cross-metric stat predicate & 0.86 & 0.97 & +0.11 & & 0.90 & 0.93 & +0.03 \\
Concordant-trend share & 0.68 & 0.92 & +0.23 & & 0.88 & 0.91 & +0.03 \\
Rank metrics by stat & 0.78 & 0.75 & $-$0.03 & & 0.78 & 0.81 & +0.03 \\
Correlation predicate & 0.68 & 0.92 & +0.23 & & 0.71 & 0.73 & +0.03 \\
Change-point detection & 0.38 & 0.48 & +0.10 & & 0.32 & 0.34 & +0.02 \\
Cluster similar metrics & 0.78 & 0.75 & $-$0.03 & & 0.69 & 0.71 & +0.02 \\
Has periodicity? & 0.82 & 0.86 & +0.04 & & 0.63 & 0.64 & +0.01 \\
Anti-corr $\times$ noise predicate & 0.68 & 0.92 & +0.23 & & 0.78 & 0.78 & +0.00 \\
Generic binary predicate & 0.77 & 0.94 & +0.17 & & 0.65 & 0.63 & $-$0.02 \\
Cross-metric stat ratio & 0.77 & 0.93 & +0.16 & & 0.88 & 0.86 & $-$0.02 \\
When does X happen? & 0.79 & 0.82 & +0.03 & & 0.89 & 0.87 & $-$0.03 \\
Cluster: anti-similar metrics & 0.78 & 0.75 & $-$0.03 & & 0.75 & 0.72 & $-$0.03 \\
Filter metrics by trend & 0.68 & 0.92 & +0.23 & & 0.86 & 0.82 & $-$0.05 \\
Trend duration share & 0.34 & 0.40 & +0.06 & & 0.70 & 0.65 & $-$0.05 \\
Enumerate trend transitions & 0.34 & 0.40 & +0.06 & & 0.79 & 0.73 & $-$0.06 \\
Enumerate trend segments & 0.34 & 0.40 & +0.06 & & 0.85 & 0.77 & $-$0.08 \\
Dominant trend type & 0.34 & 0.40 & +0.06 & & 0.74 & 0.64 & $-$0.10 \\
Events $\cap$ segments & 0.48 & 0.60 & +0.12 & & 0.64 & 0.55 & $-$0.10 \\
\midrule
Bucket mean & 0.59 & 0.70 & +0.11 & & 0.69 & 0.72 & +0.04 \\
\bottomrule
\end{tabular}
\end{adjustbox}
\end{table*}

\begin{table*}[t]
\centering
\footnotesize
\setlength{\tabcolsep}{6pt}
\caption{Compositional reasoning in Comp-R QAs. Large $\Delta_c$ values reflect compositional skills RL acquires without any CoT supervision, since the composition itself was never demonstrated during SFT.}
\label{tab:comp-id-hard}
\begin{adjustbox}{max width=\textwidth,center}
\begin{tabular}{@{}l ccc c ccc@{}}
\toprule
\multirow{2}{*}{Task} & \multicolumn{3}{c}{Atomic skills} & & \multicolumn{3}{c}{Composition} \\
\cmidrule(lr){2-4} \cmidrule(lr){6-8}
 & \oursSFT & \oursRL & $\Delta_a$ & & \oursSFT & \oursRL & $\Delta_c$ \\
\midrule
Segment-at-pos duration & 0.34 & 0.40 & +0.06 & & 0.24 & 0.86 & +0.62 \\
Trend type at position & 0.34 & 0.40 & +0.06 & & 0.32 & 0.86 & +0.54 \\
Dominant trend type & 0.34 & 0.40 & +0.06 & & 0.45 & 0.91 & +0.45 \\
Segment count & 0.34 & 0.40 & +0.06 & & 0.31 & 0.66 & +0.34 \\
Cluster count & 0.78 & 0.75 & $-$0.03 & & 0.50 & 0.80 & +0.30 \\
Amplitude vs range ratio & 0.67 & 0.71 & +0.04 & & 0.48 & 0.76 & +0.28 \\
Normalized range & 0.65 & 0.71 & +0.06 & & 0.51 & 0.78 & +0.27 \\
Asymmetric cross-metric & 0.68 & 0.92 & +0.23 & & 0.47 & 0.73 & +0.25 \\
Max in first half? & 0.79 & 0.82 & +0.03 & & 0.60 & 0.85 & +0.25 \\
Dominant cluster & 0.78 & 0.75 & $-$0.03 & & 0.40 & 0.62 & +0.23 \\
Half-mean difference & 0.70 & 0.83 & +0.13 & & 0.47 & 0.68 & +0.21 \\
Event count & 0.62 & 0.79 & +0.18 & & 0.60 & 0.80 & +0.20 \\
Concordant pair count & 0.68 & 0.92 & +0.23 & & 0.61 & 0.81 & +0.19 \\
Has periodicity? & 0.82 & 0.86 & +0.04 & & 0.61 & 0.81 & +0.19 \\
Period estimate & 0.82 & 0.86 & +0.04 & & 0.55 & 0.74 & +0.18 \\
Event near extremum & 0.70 & 0.81 & +0.11 & & 0.57 & 0.75 & +0.18 \\
Min,max in same half? & 0.77 & 0.85 & +0.08 & & 0.58 & 0.75 & +0.16 \\
Conditional cross-stat & 0.86 & 0.97 & +0.11 & & 0.44 & 0.59 & +0.16 \\
Event-recovery check & 0.62 & 0.79 & +0.18 & & 0.52 & 0.66 & +0.14 \\
Median$\approx$mean predicate & 0.79 & 1.00 & +0.21 & & 0.52 & 0.66 & +0.14 \\
Range (max$-$min) & 0.59 & 0.63 & +0.04 & & 0.68 & 0.81 & +0.13 \\
Mean shift ($\times$2 intervals) & 0.70 & 0.83 & +0.13 & & 0.54 & 0.64 & +0.10 \\
Cycle count & 0.82 & 0.86 & +0.04 & & 0.55 & 0.64 & +0.08 \\
Max-amplitude event & 0.62 & 0.79 & +0.18 & & 0.58 & 0.65 & +0.07 \\
Event type at position & 0.62 & 0.79 & +0.18 & & 0.67 & 0.73 & +0.05 \\
Cross-metric correlation est. & 0.68 & 0.92 & +0.23 & & 0.75 & 0.80 & +0.05 \\
Interval-mean compare & 0.70 & 0.83 & +0.13 & & 0.80 & 0.81 & +0.01 \\
Longest segment duration & 0.34 & 0.40 & +0.06 & & 0.83 & 0.83 & +0.00 \\
Cross-metric period compare & 0.82 & 0.86 & +0.04 & & 0.64 & 0.64 & +0.00 \\
Cross-metric event sync & 0.62 & 0.79 & +0.18 & & 0.66 & 0.66 & +0.00 \\
Mean stability ($\times$2) & 0.56 & 0.68 & +0.13 & & 0.59 & 0.58 & $-$0.01 \\
Segment-mean compare & 0.52 & 0.61 & +0.10 & & 0.67 & 0.62 & $-$0.05 \\
\midrule
Bucket mean & 0.65 & 0.75 & +0.10 & & 0.55 & 0.73 & +0.18 \\
\bottomrule
\end{tabular}
\end{adjustbox}
\end{table*}

\begin{table*}[t]
\centering
\footnotesize
\setlength{\tabcolsep}{6pt}
\caption{Compositional reasoning in OOD composite QAs. Compositions are held out at every training phase; $\Delta_c$ measures how much of the procedure RL learns transfers to unseen composition shapes.}
\label{tab:comp-ood}
\begin{adjustbox}{max width=\textwidth,center}
\begin{tabular}{@{}l ccc c ccc@{}}
\toprule
\multirow{2}{*}{Task} & \multicolumn{3}{c}{Atomic skills} & & \multicolumn{3}{c}{Composition} \\
\cmidrule(lr){2-4} \cmidrule(lr){6-8}
 & \oursSFT & \oursRL & $\Delta_a$ & & \oursSFT & \oursRL & $\Delta_c$ \\
\midrule
Concordant mean shift & 0.69 & 0.87 & +0.18 & & 0.39 & 0.76 & +0.38 \\
Mixed corr/anti-corr count & 0.59 & 0.81 & +0.23 & & 0.30 & 0.60 & +0.30 \\
Dominant trend in cluster & 0.56 & 0.57 & +0.02 & & 0.41 & 0.67 & +0.25 \\
Max before min? & 0.77 & 0.85 & +0.08 & & 0.49 & 0.72 & +0.22 \\
Mean around event & 0.70 & 0.90 & +0.19 & & 0.29 & 0.48 & +0.18 \\
Duration-weighted mean & 0.52 & 0.61 & +0.10 & & 0.32 & 0.50 & +0.18 \\
Event amp. vs seg. std & 0.65 & 0.71 & +0.06 & & 0.70 & 0.86 & +0.16 \\
Cross-corr pair count & 0.59 & 0.81 & +0.23 & & 0.38 & 0.54 & +0.16 \\
Cross-metric causality & 0.62 & 0.79 & +0.18 & & 0.51 & 0.65 & +0.14 \\
Cross-metric range overlap & 0.59 & 0.63 & +0.04 & & 0.65 & 0.78 & +0.13 \\
Quarter-mean ordering & 0.70 & 0.83 & +0.13 & & 0.67 & 0.79 & +0.13 \\
Conditional event count & 0.62 & 0.79 & +0.18 & & 0.47 & 0.59 & +0.13 \\
Segment-stat compare & 0.70 & 0.83 & +0.13 & & 0.61 & 0.74 & +0.12 \\
Cycle-mean monotone & 0.58 & 0.63 & +0.05 & & 0.64 & 0.76 & +0.12 \\
Mean conditional on trend & 0.52 & 0.61 & +0.10 & & 0.25 & 0.36 & +0.11 \\
Symmetric event recovery & 0.62 & 0.79 & +0.18 & & 0.57 & 0.66 & +0.09 \\
Event density by trend & 0.48 & 0.60 & +0.12 & & 0.36 & 0.44 & +0.08 \\
Singleton cluster check & 0.78 & 0.75 & $-$0.03 & & 0.61 & 0.69 & +0.07 \\
Correlation transitivity & 0.68 & 0.92 & +0.23 & & 0.64 & 0.71 & +0.07 \\
Position of top-mean chunk & 0.07 & 0.14 & +0.06 & & 0.48 & 0.55 & +0.07 \\
Max amp. in longest seg. & 0.47 & 0.53 & +0.05 & & 0.29 & 0.36 & +0.06 \\
Extrema alignment & 0.79 & 0.82 & +0.03 & & 0.53 & 0.59 & +0.06 \\
Chunks-above-mean fraction & 0.07 & 0.14 & +0.06 & & 0.64 & 0.70 & +0.06 \\
Symmetric trend pattern & 0.34 & 0.40 & +0.06 & & 0.54 & 0.59 & +0.05 \\
Max in top-mean quarter & 0.74 & 0.82 & +0.08 & & 0.59 & 0.63 & +0.04 \\
Std vs half-range & 0.65 & 0.71 & +0.06 & & 0.66 & 0.68 & +0.03 \\
Trend follows mean & 0.52 & 0.61 & +0.10 & & 0.56 & 0.58 & +0.02 \\
Longest-type fraction & 0.34 & 0.40 & +0.06 & & 0.66 & 0.67 & +0.01 \\
Peak in longest segment & 0.56 & 0.61 & +0.05 & & 0.52 & 0.52 & +0.00 \\
Range-normalized amplitude & 0.59 & 0.63 & +0.04 & & 0.57 & 0.54 & $-$0.02 \\
Event density & 0.62 & 0.79 & +0.18 & & 0.33 & 0.30 & $-$0.04 \\
Event amplitude vs std & 0.68 & 0.82 & +0.14 & & 0.73 & 0.69 & $-$0.04 \\
Trend reversal & 0.34 & 0.40 & +0.06 & & 0.64 & 0.50 & $-$0.13 \\
Trend convergence & 0.68 & 0.92 & +0.23 & & 0.78 & 0.59 & $-$0.19 \\
\midrule
Bucket mean & 0.57 & 0.68 & +0.11 & & 0.52 & 0.61 & +0.09 \\
\bottomrule
\end{tabular}
\end{adjustbox}
\end{table*}

\end{document}